\documentclass[10pt,twocolumn,letterpaper]{article}

\usepackage{cvpr}              

\usepackage[utf8]{inputenc} 
\usepackage[T1]{fontenc}    
\usepackage{url}            
\usepackage[english]{babel}
\usepackage{booktabs}       
\usepackage{multicol,multirow}
\usepackage[dvipsnames,table]{xcolor}
\usepackage{colortbl}
\usepackage{caption}
\usepackage{float}
\usepackage{enumitem}
\usepackage{mathrsfs}
\usepackage{savesym}
\let\checkmark\relax
\usepackage{dingbat}   
\usepackage[misc]{ifsym}
\usepackage{amssymb, amsmath, amsfonts, bm}
\usepackage{url}
\usepackage{microtype}
\usepackage{graphicx}
\usepackage{booktabs}

\usepackage{amssymb}
\usepackage{mathtools}
\usepackage{amsthm}
\usepackage{algorithm, algorithmic}
\usepackage{color, colortbl}
\definecolor{Gray}{gray}{0.9}
\usepackage{enumitem}
\usepackage{multirow}
\usepackage{enumitem, bm, multirow, array, pifont, 
            xspace, colortbl, booktabs, placeins, soul, makecell, multicol, wrapfig,
            nicefrac, scalerel, pgfplots, tikz, marginnote, subcaption, outlines, capt-of, hhline}
\pgfplotsset{compat=1.17} 
\usepackage[pagebackref,breaklinks,colorlinks,citecolor=cvprblue]{hyperref}
\usepackage{color}
\usepackage{alltt}
\usepackage{bbm}
\usepackage{graphicx}
\usepackage[capitalize]{cleveref}
\usepackage{nicefrac}
\usepackage{microtype}
\usepackage{lipsum}
\usepackage{tcolorbox}
\usepackage{diagbox}
\usepackage{tabularx}
\usepackage{kotex}
\usepackage{arydshln}

\definecolor{lightgray}{gray}{0.95}
\definecolor{cvprblue}{rgb}{0.21,0.49,0.74}

\definecolor{coco1}{HTML}{D9E4EC}
\definecolor{coco2}{HTML}{B7CFDC}
\definecolor{coco3}{HTML}{6AABD2}
\definecolor{coco4}{HTML}{385E72}
\hypersetup{
    colorlinks=true,
    linkcolor=coco3,
    filecolor=coco4,      
    urlcolor=coco3,
    citecolor=coco3,
}

\newcommand{\xmark}{\ding{55}} 

\usepackage{fontawesome}

\definecolor{cvprblue}{rgb}{0.21,0.49,0.74}

\def\paperID{3828} 
\def\confName{CVPR}
\def\confYear{2026}

\title{Multimodal Distribution Matching for Vision-Language Dataset Distillation}

\author{Jongoh Jeong\thanks{Equal contribution}\footnotemark[1]\quad Hoyong Kwon\footnotemark[1]\quad Minseok Kim\footnotemark[1]\quad Kuk-Jin Yoon\\
Visual Intelligence Lab., KAIST \\
{\tt\small \{jeong2, kwonhoyong3, alstjrx1x1, kjyoon\}@kaist.ac.kr}
\;\;\hfill
\href{https://andyj1.github.io/mdm}{\textcolor{black}{\small \faGithub\;Project Page}}
}

\begin{document}
\maketitle
\begin{abstract}
Dataset distillation compresses large training sets into compact synthetic datasets while preserving downstream performance. As modern systems increasingly operate on paired vision–language inputs, multimodal distillation must preserve representation quality and cross-modal alignment under tight compute and memory budgets, yet prior methods often require heavy computes and overlook their correlations. To address this, we present \textbf{Multimodal Distribution Matching}~(MDM), a geometry-aware framework for efficient and generalizable multimodal distillation. Specifically, MDM integrates complementary components at the data, model, and loss levels. At the data level, it initializes synthetic image–text pairs by sampling from clusters in the joint embedding space. 
At the model level, it forms a mixed teacher by interpolating independently fine-tuned models in weight space according to their angular deviation from the pretrained anchor. 
At the loss level, it matches joint distributions on the unit hypersphere using a geometry-aware matching objective that exploits the joint features in the cross-modal agreement and discrepancy directions along with symmetric contrastive learning. 
Across image–text retrieval benchmarks with cross-architecture evaluation, MDM yields compact synthetic sets that preserve multimodal semantics, substantially reduce distillation cost, and remain robust across architectures.
\end{abstract}

\section{Introduction}
\label{sec:intro}

The rapid expansion of multimodal datasets~\cite{kakaobrain2022coyo-700m, laion-5b, yfcc100m, COCO, hodosh2013flickr8k, young2014flickr} has exposed a persistent scalability barrier in modern machine learning. 
As modern vision–language models~\cite{blip, siglip, flamingo, llava, qwen, gpt4, gemini1_5} scale to hundreds of millions of paired samples, training over such collections increasingly strains compute, storage, and iteration speed for both research and deployment. This challenge is amplified in multi-modal settings, where each modality introduces its own statistical complexity and where maintaining cross-modal alignment imposes additional computational overhead. These trends underscore the growing need for compact yet informative surrogates that can preserve multi-modal semantics without incurring the full cost of large-scale training.

Dataset distillation~(DD)~\cite{wang2018datasetdistillation, zhang2024m3d, deng2024iid, cui2025optical, liu2025wmdd, li2025deda, shin2023frequency, son2024fyi, bao2025ddruo, su2024d4m, zou2025vlcp, wang2025cao2} offers a principled countermeasure by synthesizing small, representative datasets that preserve essential learning signals while cutting storage and training costs by orders of magnitude. Beyond efficiency, distillation improves reproducibility, strengthens privacy and governance by abstracting sensitive data into shareable surrogates, accelerates experimentation through faster ablations and hyperparameter sweeps, and enables edge deployment under tight memory and energy budgets. 
In vision domain, progress has advanced from gradient and feature matching to trajectory matching and generative synthesis, steadily increasing fidelity while shrinking data footprints~\cite{lei2023ddsurvey1,yu2023ddsurvey2,dd_survey25}.

\begin{figure}[!t]
    \centering
    \includegraphics[width=.99\linewidth]{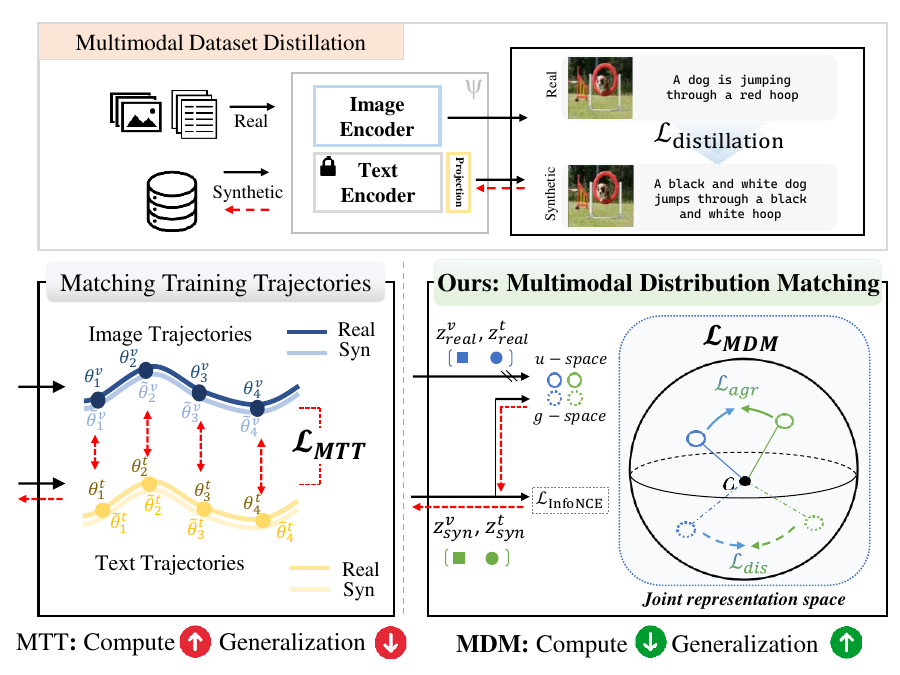}\vspace{-3mm}
    \caption{Comparison between prior multimodal dataset distillation based on matching training trajectories (MTT, \textit{left}) and our\textbf{ Multimodal Distribution Matching} (MDM, \textit{right}). While MTT replays image–text trajectories at high compute and storage cost, MDM directly matches the joint image–text distribution in the joint embedding space, yielding compact synthetic data with strong cross-architecture generalization under much lower distillation cost. The red arrow indicates the direction of gradient backpropagation.}
    \label{fig:teaser}
    \vspace{-10pt}
\end{figure}

Practical systems now widely operate on paired vision and language inputs rather than images alone, making multimodal distillation~(MDD) essential. A compact synthetic set in this regime must preserve intra-modal statistics for images and text while maintaining inter-modal semantic correspondences. 
However, compared to unimodal distillation, MDD is substantially more challenging as multimodal datasets exhibit broader semantic diversity, and the joint embedding space must simultaneously capture modality-shared and -specific information.

Existing multimodal approaches~\cite{wu2024vldistill, xu2024lors} often replay resource-intensive training trajectories or restrict optimization to narrow projection subspaces.
Trajectory-based methods~\cite{MTT, ATT, du2023ftd, zhong2025mct} repeatedly perform bi-level optimization, generating teacher and student trajectories from real and synthetic data and using them to update the synthetic set, which leads to substantial computational and memory overhead.
Moreover, optimizing synthetic data to match the training dynamics specific to the training architecture leads to architectural bias, hampering cross-architecture generalization.

These trends motivate an efficient, geometry-aware formulation that directly matches distributions in a space aligned with modern encoders. Distribution matching (DM)~\cite{DM, IDM, zhang2024dance} avoids trajectory replay and instead focuses on aligning real and synthetic feature distributions, offering advantages in scalability, stability, and generalization. Extending this idea to multi-modal learning allows us to preserve representation quality and cross-modal alignment, enabling synthetic data that better represent the underlying multi-modal distribution while remaining less sensitive to any single model’s structural bias and reducing computational load.


In this light, we present \textbf{Multimodal Distribution Matching}~(MDM), a geometry-aware distribution matching framework for MDD that performs competitively with previous methods while requiring significantly less compute. 
Our MDD framework, as shown in Fig.~\ref{fig:teaser}, effectively leverages alignment across images and text, thereby reducing sensitivity to a single set of encoder architectures. 
To this end, MDM integrates three complementary components across data, model, and distillation loss. 
Firstly, synthetic data are initialized with a broad coverage of diverse semantic modes in the image-text joint embedding space, rather than random sampling or relying on a single modality.
%
Further, MDM utilizes an image-text model whose weights are adaptively interpolated with multiple finetuned experts, operating in a more architecture-agnostic representation space to enhance cross-architecture generalization.
%
Finally, MDM optimizes synthetic data by aligning real and synthetic distributions via geodesic kernel energies over cross-modal agreement and discrepancy directions, while preserving image–text alignment within synthetic pairs via a contrastive objective.
%


\noindent We summarize our key contributions as follows:
\begin{itemize}[leftmargin=*]\setlength\itemsep{0.25em}
\item We introduce a geometry-aware MDD framework that matches real and synthetic distributions in the joint image–text embedding space, significantly reducing the distillation cost compared to previous trajectory-based methods.
\item We highlight the importance of initialization at both the data and model levels, employing joint-space data seeding and adaptive weight-space interpolation to obtain synthetic data that generalizes across architectures.
\item We design a multi-modal objective that aligns real and synthetic data by matching agreement and discrepancy directions via geodesic kernel energies, while contrastively preserving image–text alignment within synthetic pairs.
\end{itemize}
\vspace{-3pt}
\section{Related Work}\label{sec:relatedwork}

\begin{figure*}[!t]
    \centering
    \includegraphics[width=0.99\linewidth,height=9cm]{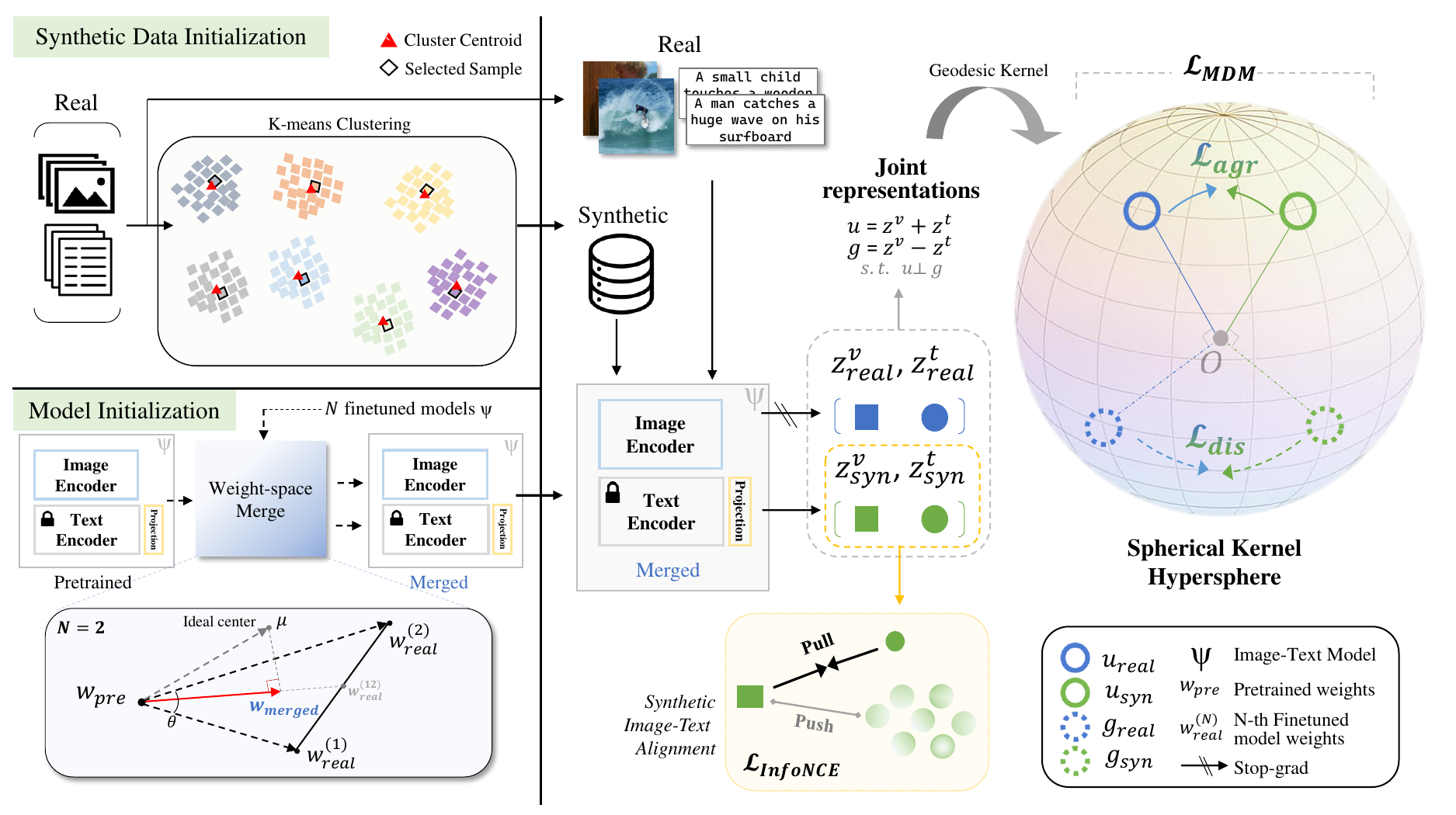}
    \vspace{-10pt}
    \caption{\textbf{Overview of MDM.} Our MDM method consists of (i) synthetic data initialization using k-means clustering, (ii) image-text model initialization using weight-space interpolation between a pretrained and $N$ finetuned models, and (iii) multimodal distribution matching that minimizes geodesic kernel energy between real and synthetic pairs on the unit hypersphere.}
    \label{fig:overview}
    \vspace{-10pt}
\end{figure*}

\paragraph{Coreset selection} aims to construct a compact subset of real training examples to retain downstream utility by emphasizing coverage or influence.
Geometry-based methods promote diversity in the representation space, including herding~\cite{c-herd} and k-center~\cite{c-kcenter}, and facility-location objectives. At the same time, dynamics-driven approaches leverage learning signals such as example forgetting~\cite{c-forget}. Probabilistic formulations construct Bayesian pseudo-coresets using variational or divergence-based criteria~\cite{manousakas2020bayesian,kim2022divergence,tiwary2023constructing}.
While these strategies perform reasonably well under moderate budgets, they operate purely by reweighting or selecting existing samples and thus are ill-suited for capturing cross-modal structures. Our approach, by contrast, synthesizes paired samples and aligns their distributions in an encoder-aligned hyperspherical feature space, enabling stronger compression while explicitly maintaining multimodal alignment.

\vspace{-8pt}
\paragraph{Dataset distillation} synthesizes compact training sets that reproduce optimization signals and generalization, and has diversified into several families. Early approaches match gradients or features between real and synthetic data~\cite{DC,DSA,FRePo,RFAD}, while trajectory matching methods align full parameter-update paths with improved scalability~\cite{MTT,cui2023tesla}.
Distribution matching aligns statistics of learned features via maximum mean discrepancy, moment matching, optimal transport, HSIC, or covariance alignment, and has shown strong cross-architecture generalization~\cite{CAFE,DM,IDM,zhang2024dance}. Other lines compress supervision into compact parameter or memory representations~\cite{IDC,linba,haba,KFS}, or frame distillation through kernels, meta-learning, implicit gradients~\cite{KIP,KIP2,RCIG}. Generative approaches directly synthesize informative samples~\cite{Generative,IT-GAN}, and recent scaling studies improve robustness and efficiency via model augmentation, slimmable architectures, pruning–recovery mechanisms, frequency cues, and representative matching~\cite{ModelAug, liu2023slimmable, yoco, sre2l, shin2023frequency, DREAM}.
In this lineage, our method contributes a compute-efficient, geometry-aware MDM scheme, leveraging joint representations as alignment and discrepancy components, and cluster-based data seeding along with angle-guided model weight interpolation for strong generalization ability of the distilled data.

\vspace{-8pt}
\paragraph{Vision–language dataset distillation} adapts these ideas to paired image–text data, where labels are implicit and the evaluation emphasizes cross-modal alignment.
The first multimodal formulation~\cite{wu2024vldistill} extends trajectory matching to contrastive retrieval training, establishing protocols on Flickr30K~\cite{young2014flickr} and COCO~\cite{COCO}. LoRS~\cite{xu2024lors} later preserves image–text similarity structure by distilling a ground-truth similarity matrix via low-rank factorization, improving efficiency in correspondence learning. While both approaches improve fidelity through dynamics-aware replay or similarity preservation, they require heavy computation and tend to bias optimization toward projection subspaces tied to specific encoder architectures, limiting cross-architecture generalization. In contrast, our method replaces trajectory and similarity-matrix supervision with distribution matching, operating on the unit hypersphere to match joint agreement and discrepancy features between real and synthetic data, thereby addressing scalability and generalization with a significantly lower computational budget.

\section{Proposed Method}\label{sec:method}
\subsection{Preliminaries} \label{sec:preliminaries}



\noindent\textbf{Problem Definition.}
Let \(\mathcal{D}_{\mathrm{real}} = \{(x_i,t_i)\}_{i=1}^{B}\) denote a real multimodal dataset of images \(x\) and texts \(t\). Our goal is to construct a much smaller synthetic dataset $\mathcal{D}_{\mathrm{syn}} = \{(\tilde x_j,\tilde t_j)\}_{j=1}^{\tilde B}$ with $| \mathcal{D}_{syn}| \ll |\mathcal{D}_{real}|$ such that models trained on $\mathcal{D}_{\mathrm{syn}}$ closely mimic the behavior of models trained on $\mathcal{D}_{\mathrm{real}}$ when evaluated on the real data distribution. 
In the vision domain, previous optimization-oriented work has sought to update $\mathcal{D}_{syn}$ using a meta-learning framework or matching model parameter weights or gradients derived from $\mathcal{D}_{syn}$ and $\mathcal{D}_{real}$. Extending to vision and language, \cite{wu2024vldistill} and \cite{xu2024lors} build on this matching trajectory regime to distill image-text data. However, the requirement of intensive computations induced by nested gradient calculations remains as a significant challenge. Distribution Matching (DM)~\cite{DM, IDM} addresses this distillation efficiency issue by directly aligning feature distributions of $\mathcal{D}_{syn}$ and $\mathcal{D}_{real}$, optimizing the following given network parameters $\theta_{0}$:
\begin{equation}
\displaystyle
\mathcal{D}_{syn}^{\star} = \operatorname*{arg\,min}_{\mathbb{E}_{\theta_{0} \sim P_{\theta_{0}}}} \Big\Vert
\frac{\textstyle\sum_{i=1}^{B} \theta_{0}\left(x_i\right)}{|\mathcal{D}_{real}|}
- 
\frac{\textstyle\sum_{j=1}^{\tilde{B}} \theta_{0}\left(\tilde{x}_j\right)}{|\mathcal{D}_{syn}|} \Big\Vert^{2}.
\end{equation}
In the same vein for efficient distillation, we extend DM to a multimodal setting to optimize $\mathcal{D}_{syn}^{\star}$ by explicitly aligning \textit{joint} image-text representations in a shared latent space as:
\begin{equation}
\label{eq:mdm_concept}
\resizebox{\linewidth}{!}{$
\mathcal{D}_{\mathrm{syn}}^{\star} =
\displaystyle\operatorname*{arg\,min}_{\mathcal{D}_{\mathrm{syn}}}
\phi\Big(
\underbrace{
\mathbb{E}_{(X,T)\sim\mathcal{D}_{real}}
\big[\Psi(X,T)\big]
}_{\text{real joint}}
,
\underbrace{
\mathbb{E}_{(\tilde{X},\tilde{T})\sim\mathcal{D}_{syn}}
\big[\Psi(\tilde{X},\tilde{T})\big]
}_{\text{synthetic joint}}
\Big),
$}
\end{equation}
\noindent where $\Psi(\cdot, \cdot)$ denotes a joint feature extractor comprising an image $\theta^{v}$ and a text $\theta^{t}$ encoder with projection, and $\phi(\cdot, \cdot)$ is a distance function.
See \textit{Supp.} for detailed formulation.

\noindent\textbf{Image-Text Matching.}
Image-text retrieval is a central probe for vision--language understanding, and we thus adopt the commonly used bidirectional InfoNCE objective~\cite{radford2021clip} as the base pairwise alignment loss for the synthetic data. 
Given a minibatch $\{(\tilde x_j,\tilde t_j)\}_{j=1}^{\tilde B} \subset \mathcal{D}_{\mathrm{syn}}$, we obtain image features $z^v_j$ and text features $z^t_j$ from the image and text branches of the unified image-text model $\Psi$. 
We then define similarity logits
$\tilde{Z}_{jk}=\mathbf{s}(\tilde{z}^v_j, \tilde{z}^t_k)/\tau$ using a chosen similarity function $\mathbf{s}(\cdot,\cdot)$ (e.g., cosine or dot product) and temperature $\tau>0$. 
The InfoNCE loss averages the image$\rightarrow$text and text$\rightarrow$image cross-entropy terms with identity targets, instantiated for the synthetic image-text features ($\tilde{z}^{v}, \tilde{z}^{t}$) as:
\vspace{-2mm}
\begin{equation}
\label{eq:loss_infonce}
\resizebox{\linewidth}{!}{$
\mathcal{L}_{\mathrm{InfoNCE}}
:=\frac{1}{2\tilde B}
\displaystyle\sum_{j=1}^{\tilde B}\!
\left[
\underbrace{
 -\log \frac{\exp(\tilde{Z}_{jj})}{\sum_{k=1}^{\tilde B}\exp(\tilde{Z}_{jk})}
}_{\text{$v \rightarrow t$}}
+
\underbrace{
 -\log \frac{\exp(\tilde{Z}_{jj})}{\sum_{k=1}^{\tilde B}\exp(\tilde{Z}_{kj})}
}_{\text{$t \rightarrow v$}}
\right],
$}
\end{equation}
which encourages each synthetic image $\tilde x_j$ to rank its paired synthetic caption $\tilde t_j$ above all others in the batch, and symmetrically, each caption to rank its paired image above all others. 
In our MDM formulation in Eq.~\ref{eq:mdm_concept}, this contrastive objective acts as the alignment component operating on the synthetic pairs, while the distribution matching term aligns their joint representations with those of the real data.

\subsection{Synthetic Data Initialization}
We first initialize the synthetic dataset $\mathcal{D}_{\mathrm{syn}}$ by selecting representative real samples in the joint embedding space of the encoder $\Psi$. Let $f_n $ denote the joint embedding of the $n$-th real pair, obtained by concatenating the image and text features after projection. We embed all real pairs, run $K$-means clustering~\cite{hartigan1979kmeans} with $K = |\mathcal{D}_{\mathrm{syn}}|$, and obtain cluster centroids $\{c_k\}_{k=1}^{K}$ and indices $\{\mathcal{C}_k\}_{k=1}^{K}$. For each cluster, we initialize each synthetic pair with a real sample whose joint feature is closest to its centroid in cosine distance:
\vspace{-2mm}
\begin{equation}\label{eq:kmeans_init}
\resizebox{.95\linewidth}{!}{$
\mathcal{D}_{\mathrm{syn}}^{(0)}
=
\bigl\{(x_{j_k}, t_{j_k})\bigr\}_{k=1}^{K},
\quad \mathrm{with}\;\;
j_k
=
\operatorname*{\arg\max}_{n \in \mathcal{C}_k}
\frac{f_n^\top c_k}{\|f_n\|_2 \,\|c_k\|_2}.
$}
\end{equation}
This cluster-based seeding induces synthetic data to be composed of representative samples with a broad coverage of joint semantic modes in the joint space while avoiding redundancy.  
Because $f_n$ is constructed from concatenated image and text features, the clustering reflects a multimodal structure rather than either modality alone—providing a significantly stronger initialization for distillation.  
Since the synthetic pairs are later optimized as continuous parameters, this initialization offers a stable starting point aligned with the real-data manifold. Importantly, seeding is performed entirely in the feature space but is instantiated as concrete image-text pairs, enabling architecture-agnostic initialization that complements our model initialization.

\subsection{Image-Text Model Initialization}

Beyond data initialization, the image–text model's initialization is crucial for distilling multimodal data, as it largely determines the geometry of the joint embedding space. If the model remains too close to the pretrained anchor $\theta_{0}^{\mathrm{v/t}}$, the resulting teacher may underfit the real multimodal structure, weakening the signal from the real set that synthetic data must eventually mimic. 
Conversely, leaning too heavily on a single finetuned expert causes the synthetic data to inherit that expert’s model-specific representation geometry rather than the full real-data distribution. This leads to degraded distillation performance and diminished cross-architecture generalization. We therefore shift our image-text model towards the real only to the extent that $N$ finetuned models agree in direction, using their agreement to retain the initial model. In the process, we exploit the finetuned experts used in prior work~\cite{wu2024vldistill, xu2024lors}. 

Concretely, inspired by the geometric perspective that the center-close weight approximation of finetuned models yields a significant improvement in classification robustness \cite{jang2024modelstock}, we adapt its weight-space interpolation to our image encoder and text projection modules. 
For the pretrained anchor parameters $\theta^v_{0}$ and $\theta^t_{0}$ as image encoder and text projector, respectively, and $N$ (\eg, 2) independent finetuned experts $(\theta^v_{(1)},\theta^t_{(1)})$ and $(\theta^v_{(2)},\theta^t_{(2)})$ trained on real data, we merge layer-wise ($\ell$) weights as follows:
\vspace{-1mm}
\begin{equation}\label{eq:weight_merge}
\resizebox{.95\linewidth}{!}{$
\begin{aligned}
\theta^{m}_{*,\ell}
&=
\theta^{m}_{0,\ell}
+\alpha\,t^{m}_{\ell}\cdot \tfrac{1}{2}\big(\Delta^{m}_{1,\ell}+\Delta^{m}_{2,\ell}\big),
\;\; \mathrm{for}\; m\in\{v,t\},
\\[2pt]
\textrm{with}\quad
t^{m}_{\ell}
&=
\frac{2\,\big\langle \Delta^{m}_{1,\ell},\,\Delta^{m}_{2,\ell}\big\rangle}
{\big\|\Delta^{m}_{1,\ell}\big\|_2\,\big\|\Delta^{m}_{2,\ell}\big\|_2
+\big\langle \Delta^{m}_{1,\ell},\,\Delta^{m}_{2,\ell}\big\rangle}
\end{aligned},
$}
\end{equation}

\noindent where $\langle\cdot,\cdot\rangle$ denotes the inner product. $\Delta^{m}_{i,\ell} \coloneqq \theta^{m}_{(i),\ell}-\theta^{m}_{0,\ell}$ denotes the displacement of the expert $i\in\{1,\cdots,N\}$ from the pretrained anchor in layer \(\ell\) for the modality \(m\), and $t^m_\ell$ is the merging ratio determined by the mutual angle between the displacement vectors. Here, the larger angle deviation between the finetuned experts, the greater the reliance of the merged weight on the pretrained anchor, and vise versa.
We further finely tune the level of weight shift with $\alpha<1$ as we target architectural robustness, which may be more sensitive in the weight space. This layer-wise weight interpolation step adapts only when the directions of displacements align and defaults to the anchor when they conflict, thereby yielding merged weights with strong architectural generalization.

At each distillation iteration, we apply this step after selecting $N$ checkpoints of a random expert at a random epoch from its training trajectory. As each expert follows distinct per-epoch training dynamics, refreshing the merge with a randomly sampled expert checkpoint at each iteration implicitly mimics the real-distribution dynamics at the model level rather than the parameter level, as practiced explicitly in MTT~\cite{MTT, wu2024vldistill, xu2024lors}. This encourages and stabilizes the supervision signal for $\mathcal{D}_{syn}$, and empirically strengthens cross-architecture generalization.

\subsection{Multimodal Distribution Matching}

In this section, we describe our MDM approach, which exploits the geometry of joint image-text features on a geodesic kernel~\cite{mei2025geomm, kimmel1998computing}, illustrated in Fig.~\ref{fig:overview}. See \textit{Supp.} for details.

\noindent\textbf{Notation and Joint Feature Space.}
Given a pair of image-text $(x,t)$ and its embeddings after projection $(z^v, z^t)$, \ie, the output of $\Psi(x,t)$, we utilize the normalized embeddings so that they lie on the unit hypersphere $\in \mathbb{S}^{d-1}$.
Then, for each pair $(z^v,z^t)$, we construct cross-modal features in the \emph{agreement} and \emph{discrepancy} directions on $\mathbb{S}^{d-1}$, s.t.:
\begin{equation}
u \;=\; \mathrm{normalize}(z^v + z^t),\;\;g \;=\; \mathrm{normalize}(z^v - z^t),
\end{equation}
\noindent where $\mathrm{normalize}(\cdot) := \frac{\cdot}{\|\cdot\|_2}$. For each mini-batch, we construct batch sets: 
$\mathcal{U}^{r}=\{u_i^{r}\}_{i=1}^{B^r}$, $\mathcal{U}^{s}=\{u_j^{s}\}_{j=1}^{B^s}$, $\mathcal{G}^{r}=\{g_i^{r}\}_{i=1}^{B^r}$, $\mathcal{G}^{s}=\{g_j^{s}\}_{j=1}^{B^s}$
, where $r$ denotes real set and $s$ denotes synthetic set.

\vspace{-5pt}
\paragraph{Geodesic Kernel Energy.}
With the modality embeddings $(z^v,z^t)$ and $(u,g)$ lying on the unit hypersphere $\mathbb{S}^{d-1}$, We compute the angular similarity between each real-synthetic pair of $u$'s and $g$'s as follows:
\begin{equation}
\label{eq:theta}
\phi(a,b) \;=\; \arccos\!\big(\langle a,b\rangle\big)\;\in[0,\pi],\qquad a,b\in\mathbb{S}^{d-1}.
\end{equation}
With the similarity angles, we kernelize the features with a geodesic Gaussian kernel to map to a hyperspherical affinity:
\begin{equation}
\label{eq:kgeo}
k_{\mathrm{geo}}(a,b) \;=\; \exp\!\Big(-\frac{\phi(a,b)^2}{2\,\sigma^2}\Big),\qquad \sigma>0.
\end{equation}
The kernel defines a pairwise \emph{energy} on $\mathbb{S}^{d-1}$ that compares affinities under the same intrinsic metric, yielding a common scale for relating intra- and inter-set alignment.
For two finite sets $\mathcal{A}=\{a_i\}_{i=1}^{m}$ and $\mathcal{B}=\{b_j\}_{j=1}^{n}$ on $\mathbb{S}^{d-1}$, we define the geodesic kernel energy with intra-set affinity on both sets along with inter-set affinity:
\begin{equation}
\label{eq:gke}
\resizebox{1.01\linewidth}{!}{$
\mathsf{GKE}(\mathcal{A},\mathcal{B})
\;=\;
\Bigg[
\underbrace{\frac{1}{m^2}\sum_{i=1}^{m}\sum_{i'=1}^{m}\! k_{\mathrm{geo}}(a_i,a_{i'})}_{\textrm{intra-}\mathcal{A}}
\;+\;
\underbrace{\frac{1}{n^2}\sum_{j=1}^{n}\sum_{j'=1}^{n}\! k_{\mathrm{geo}}(b_j,b_{j'})}_{\textrm{intra-}\mathcal{B}}
\;-\;
\underbrace{\frac{2}{mn}\sum_{i=1}^{m}\sum_{j=1}^{n}\! k_{\mathrm{geo}}(a_i,b_j)}_{\textrm{inter-set affinity}}
\Bigg]^{\!1/2},
$}
\end{equation}
\noindent where the square root term moderates the scale relative to the angular magnitudes and produces smoother gradients. 
Minimization increases inter-set affinity while calibrating the internal affinity patterns of the two input sets under the same geodesic kernel.
We refer to \textit{Supp.} for detailed formulation.
Given these cross-modal agreement and discrepancy vectors from the real and synthetic mini-batches, we define the geodesic kernel energy in Eq.~\ref{eq:gke_ud} as a distribution on the hyperspherical kernel to match between real and synthetic data.
\begin{equation}
\label{eq:gke_ud}
\mathcal{L}_{\mathrm{agr}} \;=\; \mathsf{GKE}\!\big(\mathcal{U}^{r},\mathcal{U}^{s}\big),
\qquad
\mathcal{L}_{\mathrm{dis}} \;=\; \mathsf{GKE}\!\big(\mathcal{G}^{r},\mathcal{G}^{s}\big).
\end{equation}

\vspace{-5pt}
\paragraph{Total Distillation Objective.}
We combine the geodesic kernel energies with a bidirectional InfoNCE term at temperature $\tau$ over a batch of $B$ paired examples to formulate our total loss as follows:
\begin{equation}
\mathcal{L}_{\textrm{MDM}} = \mathcal{L}_{\mathrm{InfoNCE}} + \lambda_{\mathrm{agr}}\cdot\mathcal{L}_{\mathrm{agr}} + \lambda_{\mathrm{dis}}\cdot\mathcal{L}_{\mathrm{dis}},
\end{equation}
\noindent where $\lambda_{agr}$ and $\lambda_{dis}$ denote tuning factors for the cross-modal agreement and discrepancy losses, respectively.

\section{Experiments}
\label{sec:experiments}

\begin{table*}[!t]
\caption{\textbf{Image-text retrieval results} for 100, 200, and 500 synthetic pairs using the coreset methods and distillation method. The condensation rate for \{Flickr8k, Flickr30k, and COCO\} datasets are approximately \{1.7\%, 0.3\%, 0.8\textperthousand\}, \{3.3\%, 0.7\%, 1.7\textperthousand\}, \{8.3\%, 1.7\%, 4.4\textperthousand\} for 100, 200, and 500 pairs. Best and runner-up results are indicated in \textbf{boldface} and \underline{underline}, respectively.}
\vspace{-2mm}
\label{tab:main}
\centering
\setlength{\tabcolsep}{2pt}
\renewcommand{\arraystretch}{1}
\renewcommand{\aboverulesep}{3pt}
\renewcommand{\belowrulesep}{3pt}
\vspace{-1mm}
\resizebox{\linewidth}{!}{%
\begin{tabular}{cl|ccccccc ccccccc ccccccc}
\toprule
\multirow{2}{*}{\rotatebox[origin=c]{90}{\# Pairs}}
& \multirow{1}{*}{\qquad\quad Dataset}
& \multicolumn{7}{c}{Flickr8k~\cite{hodosh2013flickr8k}} 
& \multicolumn{7}{c}{Flickr30k~\cite{young2014flickr}} 
& \multicolumn{7}{c}{COCO~\cite{COCO}} \\
\cmidrule(lr){3-9} \cmidrule(lr){10-16} \cmidrule(lr){17-23} 
& \multirow{1}{*}{Method}
& \textrm{IR@1} & \textrm{IR@5} & \textrm{IR@10} & \textrm{TR@1} & \textrm{TR@5} & \textrm{TR@10} & \cellcolor{Goldenrod!20} \textrm{Mean}
& \textrm{IR@1} & \textrm{IR@5} & \textrm{IR@10} & \textrm{TR@1} & \textrm{TR@5} & \textrm{TR@10} & \cellcolor{Goldenrod!20}\textrm{Mean}
& \textrm{IR@1} & \textrm{IR@5} & \textrm{IR@10} & \textrm{TR@1} & \textrm{TR@5} & \textrm{TR@10} & \cellcolor{Goldenrod!20}\textrm{Mean} \\
\midrule
\multirow{8}{*}{\rotatebox{90}{100}}
& Random 
& 1.2 & 5.6 & 9.6 & 2.7 & 8.0 & 12.6 & \cellcolor{Goldenrod!20}6.6
& 0.9 & 4.2 & 7.3 & 2.0 & 7.9 & 12.1 & \cellcolor{Goldenrod!20}5.7
& 0.4 & 1.7 & 3.0 & 0.8 & 3.2 & 5.4 & \cellcolor{Goldenrod!20}2.4 \\

& Herding~\cite{c-herd} 
& 1.2 & 4.4 & 8.5 & 2.2 & 8.5 & 14.2 & \cellcolor{Goldenrod!20}6.5
& 0.9 & 3.5 & 6.5 & 2.0 & 6.9 & 11.1 & \cellcolor{Goldenrod!20}5.1
& 0.3 & 1.4 & 2.6 & 0.8 & 3.0 & 5.5 & \cellcolor{Goldenrod!20} 2.4 \\

& K-Center~\cite{c-kcenter} 
& 1.2 & 4.9 & 9.0 & 2.7 & 9.3 & 13.9 & \cellcolor{Goldenrod!20}6.8
& 1.1 & 4.9 & 8.7 & 3.0 & 9.1 & 14.3 & \cellcolor{Goldenrod!20}6.8
& 0.5 & 1.9 & 3.6 & 1.1 & 4.2 & 7.6 & \cellcolor{Goldenrod!20}3.2 \\

& Forgetting~\cite{c-forget} 
& 1.2 & 4.0 & 7.1 & 1.5 & 4.8 & 8.4 & \cellcolor{Goldenrod!20}4.5
& 0.8 & 3.6 & 6.2 & 1.2 & 5.4 & 9.1 & \cellcolor{Goldenrod!20}4.4
& 0.2 & 1.0 & 1.9 & 0.2 & 1.2 & 2.6 & \cellcolor{Goldenrod!20}1.2 \\[0.5ex]

\cdashline{2-23}\noalign{\vskip 0.5ex}

& MTT-VL~\cite{wu2024vldistill} 
& 0.8 & 4.0 & 7.0 & 1.5 & 6.4 & 10.8 & \cellcolor{Goldenrod!20}5.1
& 4.7 & 15.7 & 24.6 & 9.9 & 28.3 & 39.1 & \cellcolor{Goldenrod!20}20.4
& 1.3 & 5.4 & 9.5 & 2.5 & 10.0 & 15.7 & \cellcolor{Goldenrod!20}7.4 \\

& TESLA$_{\textrm{WBCE}}$~\cite{cui2023tesla}
& 0.8 & 3.8 & 7.0 & 4.7 & 16.1 & 25.9 & \cellcolor{Goldenrod!20} 9.7 
& 0.5 & 2.3 & 4.7 & 5.5 & 19.5 & 28.9 & \cellcolor{Goldenrod!20}10.2
& 0.3 & 1.0 & 1.8 & 2.0 & 7.7 & 13.5 & \cellcolor{Goldenrod!20}4.4 \\

& LoRS~\cite{xu2024lors} 
& 4.9 & 18.0 & 29.0 & 7.0 & 22.8 & 34.8 & \cellcolor{Goldenrod!20}\underline{19.4}
& 8.3 & 24.1 & 35.1 & 11.8 & 35.8 & 49.2 & \cellcolor{Goldenrod!20}\textbf{27.4}
& 1.8 & 7.1 & 12.2 & 3.3 & 12.2 & 19.6 & \cellcolor{Goldenrod!20}\underline{9.4} \\

& \cellcolor{Goldenrod!20}Ours
& 6.0 & 20.8 & 32.4 & 7.9 & 26.5 & 38.1          &\cellcolor{Goldenrod!20}\textbf{21.9} 
&8.1	&24.7	&36.2&	11.5&	32.6&	45.0	& \cellcolor{Goldenrod!20}\underline{26.4}
& 1.9 &  7.6 & 13.2   &    3.6 & 13.7 &  21.6  & \cellcolor{Goldenrod!20}\textbf{10.3} \\

\midrule

\multirow{8}{*}{\rotatebox{90}{200}}
& Random 
& 2.0 & 7.8 & 13.7 & 3.3 & 12.5 & 19.5 & \cellcolor{Goldenrod!20}9.8
& 1.9 & 7.1 & 12.3 & 1.9 & 10.3 & 18.2 & \cellcolor{Goldenrod!20}8.6
& 0.6 & 2.7 & 4.9 & 1.3 & 5.3 & 9.0 & \cellcolor{Goldenrod!20}4.0 \\

& Herding~\cite{c-herd} 
& 2.0 & 7.6 & 14.0 & 3.2 & 12.5 & 19.9 & \cellcolor{Goldenrod!20}9.9
& 1.4 & 5.9 & 10.5 & 3.1 & 9.4 & 15.5 &\cellcolor{Goldenrod!20} 7.6
& 0.6 & 2.5 & 4.6 & 1.1 & 4.6 & 8.4 & \cellcolor{Goldenrod!20}3.6\\

& K-Center~\cite{c-kcenter} 
& 2.3 & 9.1 & 15.0 & 3.8 & 13.7 & 20.9 & \cellcolor{Goldenrod!20} 10.8
& 2.2 & 8.1 & 13.3 & 4.2 & 13.1 & 21.2 & \cellcolor{Goldenrod!20}10.3
& 0.9 & 3.4 & 5.9 & 2.1 & 7.0 & 11.6 & \cellcolor{Goldenrod!20}5.1 \\

& Forgetting~\cite{c-forget} 
& 1.7 & 6.5 & 11.5 & 3.1 & 9.7 & 15.4 & \cellcolor{Goldenrod!20}8.0
& 1.6 & 6.6 & 10.8 & 2.5 & 9..0 & 14.9 & \cellcolor{Goldenrod!20}7.6
& 0.4 & 1.6 & 3.0 & 0.7 & 2.8 & 5.1 & \cellcolor{Goldenrod!20}2.3 \\[0.5ex]

\cdashline{2-23}\noalign{\vskip 0.5ex}

& MTT-VL~\cite{wu2024vldistill} 
& 1.8 & 7.0 & 12.2 & 2.8 & 10.3 & 17.3 & \cellcolor{Goldenrod!20}8.6
& 4.6 & 16.0 & 25.5 & 10.2 & 28.7 & 41.9 & \cellcolor{Goldenrod!20}21.2
& 1.7 & 6.5 & 12.3 & 3.3 & 11.9 & 19.4 & \cellcolor{Goldenrod!20}9.2 \\

& TESLA$_{\textrm{WBCE}}$~\cite{cui2023tesla}
& 1.2 & 4.7 & 8.4 & 6.6 & 19.5 & 29.5 & \cellcolor{Goldenrod!20}11.7
& 0.2 & 1.3 & 2.5 & 2.8 & 10.4 & 17.4 & \cellcolor{Goldenrod!20}5.8
& 0.1 & 0.2 & 0.5 & 0.7 & 3.1 & 5.3 & \cellcolor{Goldenrod!20}1.7 \\

& LoRS~\cite{xu2024lors} 
& 6.3 & 20.5 & 31.6 & 9.5 & 26.3 & 38.2 & \cellcolor{Goldenrod!20}\underline{22.1}
& 8.6 & 25.3 & 36.6 & 14.5 & 38.7 & 53.4 & \cellcolor{Goldenrod!20}\textbf{29.5}
& 2.4 & 9.3 & 15.5 & 4.3 & 14.2 & 22.6 & \cellcolor{Goldenrod!20}\underline{11.4} \\

& \cellcolor{Goldenrod!20}Ours
& 7.1	&23.2&	35.1	&	9.9	&29.0	&41.6	& \cellcolor{Goldenrod!20}\textbf{24.3}
& 9.1&	26.7&	39.1	&	13.0&	33.7&	47.4&	\cellcolor{Goldenrod!20}\underline{28.2} 
& 2.9  &  11.1  &  18.4 &   4.9  &  16.2  &  25.3   &\cellcolor{Goldenrod!20}\textbf{13.1} \\

\midrule

\multirow{8}{*}{\rotatebox{90}{500}}
& Random 
& 3.7 & 13.0 & 21.2 & 6.0 & 19.4 & 28.8 & \cellcolor{Goldenrod!20}15.3
& 3.2 & 11.5 & 18.9 & 5.2 & 18.3 & 27.4 &\cellcolor{Goldenrod!20} 14.1
& 1.2 & 5.2 & 9.2 & 2.5 & 8.7 & 14.9 & \cellcolor{Goldenrod!20}7.0\\

& Herding~\cite{c-herd} 
& 3.7 & 12.5 & 19.8 & 4.9 & 17.5 & 26.4 & \cellcolor{Goldenrod!20}14.1
& 2.7 & 10.6 & 17.0 & 4.1 & 14.9 & 24.0 & \cellcolor{Goldenrod!20}12.1
& 1.3 & 5.0 & 8.8 & 2.0 & 7.9 & 13.6 & \cellcolor{Goldenrod!20}6.4\\

& K-Center~\cite{c-kcenter} 
& 4.0 & 13.4 & 21.1 & 5.9 & 18.9 & 29.0 & \cellcolor{Goldenrod!20}15.4
& 3.4 & 11.8 & 18.7 & 6.7 & 18.0 & 30.6 & \cellcolor{Goldenrod!20}14.9
& 1.5 & 5.7 & 9.7 & 3.0 & 9.9 & 16.2 & \cellcolor{Goldenrod!20}7.7\\

& Forgetting~\cite{c-forget} 
& 4.6 & 16.2 & 24.5 & 5.8 & 21.7 & 31.7 & \cellcolor{Goldenrod!20}17.4
& 3.6 & 12.7 & 20.6 & 6.1 & 18.7 & 29.5 & \cellcolor{Goldenrod!20}15.2
& 1.1 & 4.3 & 7.6 & 2.0 & 7.3 & 11.3 & \cellcolor{Goldenrod!20}5.6 \\[0.5ex]

\cdashline{2-23}\noalign{\vskip 0.5ex}

& MTT-VL~\cite{wu2024vldistill} 
& 3.8 & 13.3 & 21.2 & 5.7 & 18.6 & 27.7 & \cellcolor{Goldenrod!20}15.1
& 6.6 & 20.2 & 30.0 & 13.3 & 32.8 & 46.8 & \cellcolor{Goldenrod!20}25.0
& 2.5 & 8.9 & 15.8 & 5.0 & 17.2 & 26.0 & \cellcolor{Goldenrod!20}12.6 \\

&
TESLA$_{\textrm{WBCE}}$~\cite{cui2023tesla}
& 2.5 & 8.8 & 14.1 & 6.9 & 19.6 & 29.0 & \cellcolor{Goldenrod!20}13.5
& 1.1 & 7.3 & 12.6 & 5.1 & 15.3 & 23.8 & \cellcolor{Goldenrod!20}10.9
& 0.8 & 3.6 & 6.7 & 1.7 & 5.9 & 10.2 & \cellcolor{Goldenrod!20}4.8\\

&  LoRS~\cite{xu2024lors} 
& 6.9 & 22.0 & 33.1 & 10.9 & 31.0 & 45.8 & \cellcolor{Goldenrod!20}\underline{25.0}
& 10.0 & 28.9 & 41.6 & 15.5 & 29.8 & 53.7 & \cellcolor{Goldenrod!20}\textbf{31.6}
& 2.8 & 9.9 & 16.5 & 5.3 & 18.3 & 27.9 & \cellcolor{Goldenrod!20}\underline{13.5}\\
&
\cellcolor{Goldenrod!20} Ours 
& 7.4 & 25.0 & 37.1 & 11.2 & 32.4 & 44.2 &\cellcolor{Goldenrod!20}\textbf{26.2}
& 10.0	& 29.3	& 42.0 &	13.7&	37.0	&51.5&	\cellcolor{Goldenrod!20}\underline{30.6}
&3.7	& 13.6 &	22.2	&5.6&	18.4	& 28.2 & \cellcolor{Goldenrod!20}\textbf{15.3} \\

\midrule

& Full Dataset 
& 25.5 & 56.1 & 69.2 & 32.7 & 64.5 & 74.5 & \cellcolor{Goldenrod!20}53.8 
& 28.1 & 57.9 & 70.3 & 34.9 & 65.4 & 77.6 & \cellcolor{Goldenrod!20}55.7
& 17.3 & 42.8 & 56.7 & 20.4 & 47.7 & 62.1 & \cellcolor{Goldenrod!20}41.2 \\

\bottomrule
\end{tabular}%
}
\vspace{-10pt}
\end{table*}

\begin{figure*}[!t]
\centering
\vspace{-0pt}
\includegraphics[width=.99\linewidth]{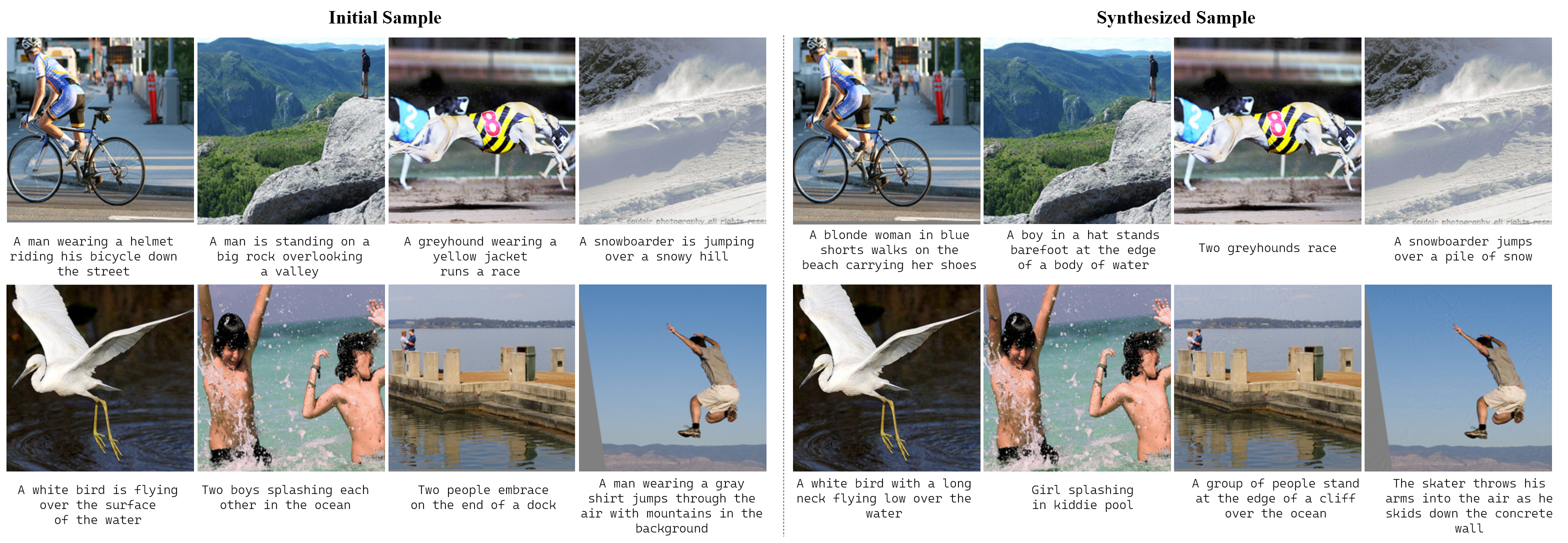}\vspace{-3mm}
\caption{\textbf{Qualitative results of synthesized data.} We compare the initial (\textit{left}) and distilled samples (\textit{right}).}
\label{fig:qualitative}
\vspace{-10pt}
\end{figure*}

\subsection{Experimental Details}

\noindent\textbf{Datasets, Tasks, and Baselines.}
We evaluate on three captioning datasets commonly used for image--text retrieval:
Flickr8k~\cite{hodosh2013flickr8k} (a total of 8,000 images, each paired with five human-written captions), Flickr30k~\cite{young2014flickr} ($\approx$31k images, five captions per image), and MS~COCO~\cite{COCO} ($\approx$123k images with five captions per image). Following standard practice on the Karpathy splits, we report retrieval performance for both text-to-image (T$\rightarrow$I) and image-to-text (I$\rightarrow$T) using Recall at \(K\in\{1,5,10\}\) (\ie, R@1/5/10). For instance, IR@5 denotes text-to-image recall at K=5, while TR@1 denotes image-to-text recall at K=1. Unless otherwise noted, COCO evaluation uses the 5k-image test set. For comparison, we include two families of strong baselines: (i) Coreset selection: Random sampling, Herding~\cite{c-herd}, K-Center~\cite{c-kcenter}, and Forgetting~\cite{c-forget}, and (ii) dataset distillation: MTT-VL~\cite{wu2024vldistill}, TESLA~\cite{cui2023tesla}, and LoRS~\cite{xu2024lors}. We adopt the publicly available configurations to re-implement the baselines for fair comparisons (see details in \textit{Supp.}).

\noindent\textbf{Architectures.}
Following MTT-VL~\cite{wu2024vldistill} and LoRS~\cite{xu2024lors}, we employ Normalizer-Free Network (NFNet)~\cite{brock2021nfnet} as the vision encoder backbone that dispenses with batch normalization in favor of normalization-free residual blocks stabilized by scaled weight standardization and careful initialization. The final convolutional features are global-average pooled and passed through a lightweight projection head to a \(d\)-dimensional embedding, then \(\ell_2\)-normalized. 
For the text side, we use BERT~\cite{devlin2018bert}. The text embedding is then linearly projected onto the same \(d\)-dimensional space and \(\ell_2\)-normalized. Image and text embeddings are trained under a contrastive alignment objective. In inference, retrieval uses cosine similarity in the joint space.

\noindent\textbf{Implementation Details.}
We distill synthetic image–text queries of size $N\in\{100,200,500\}$ using pretrained vision and text encoders assembled via parameter interpolation of per-epoch buffered checkpoints. At each distillation iteration, the frozen encoder is re-initialized using an interpolation factor $\alpha=0.5$. Following prior work~\cite{wu2024vldistill, xu2024lors}, we distill text embeddings rather than raw captions, jointly optimizing $N$ synthetic images of size $3\times224\times224$ and 768-dimensional text embeddings. Distillation is performed for up to 3,000 iterations using SGD with a fixed learning rate, momentum 0.5, and gradient clipping 1.0, where convergence typically occurs much earlier in practice. We use temperature $\tau=0.07$ and weighting coefficients $\lambda_{agr}=0.8$, $\lambda_{dis}=0.8$. After distillation, each initialization model is trained for 100 epochs on the synthetic pairs and evaluated on the real test set; all reported retrieval scores are averaged over five independently initialized models. 
See \textit{Supp.} for additional implementation details. 
%
%


\subsection{Main Results}\label{sec:main_exp}
\vspace{-4pt}
\noindent\textbf{Image-Text Retrieval.}
We report in Table~\ref{tab:main} the image–text retrieval performance in all three datasets of different scales—Flickr8k~\cite{hodosh2013flickr8k}, Flickr30k~\cite{young2014flickr}, and COCO (123k)~\cite{COCO}—given distillation budgets of 100, 200 and 500 synthetic pairs. We distill synthetic data using NFNet as thet vision encoder and BERT as the text encoder to ensure a fair comparison with the baselines~\cite{wu2024vldistill, xu2024lors}.
Across all datasets and budget regimes, our method consistently outperforms the trajectory-matching baselines MTT-VL~\cite{wu2024vldistill} and TESLA$_{\textrm{WBCE}}$~\cite{cui2023tesla}, achieving superior scores over all retrieval metrics. 
Compared to the current MTT-based state-of-the-art, LoRS~\cite{xu2024lors}, our MDM achieves competitive performance across the board. 
Notably, in the largest and most challenging COCO dataset, our method outperforms LoRS by a noticeable margin, demonstrating that MDM maintains robustness even with a minimal distillation budget (condensation rate). 
Overall, these results demonstrate the effectiveness and scalability of our MDM framework, even for challenging low-budget distillation scenarios.

\begin{table*}[!t]
\caption{\textbf{Cross-architecture generalization}. We report the averaged results over retrieval metrics including IR/TR@K=\{1,5,10\}. Note that the source model results denoted with `$^{\ast}$' are not averaged, and the best results are in \textbf{boldface}. (a)--(c): NFNet, NF-ResNet, NF-RegNet.}
\vspace{-2mm}
\label{tab:cross}
\centering
\setlength{\tabcolsep}{3pt}
\renewcommand{\arraystretch}{1}
\renewcommand{\aboverulesep}{3pt}
\renewcommand{\belowrulesep}{3pt}
\vspace{-1mm}
\resizebox{\linewidth}{!}{%
\begin{tabular}{cr cccccccc c cccccccc c cccccccc}
\toprule
&& \multicolumn{8}{c}{Flickr8k~\cite{hodosh2013flickr8k}} && \multicolumn{8}{c}{Flickr30k~\cite{young2014flickr}} && \multicolumn{8}{c}{COCO~\cite{COCO}} \\
\cmidrule{3-10}\cmidrule{12-19}\cmidrule{21-28}
\multirow{2}{*}{\rotatebox[origin=c]{90}{\# Pairs}}
& Text
& \multicolumn{3}{c}{BERT~\cite{devlin2018bert}} 
&& \multicolumn{3}{c}{DistilBERT~\cite{sanh2019distilbert}} & 
&& \multicolumn{3}{c}{BERT~\cite{devlin2018bert}} 
&& \multicolumn{3}{c}{DistilBERT~\cite{sanh2019distilbert}} & 
&& \multicolumn{3}{c}{BERT~\cite{devlin2018bert}} 
&& \multicolumn{3}{c}{DistilBERT~\cite{sanh2019distilbert}} & \\ 
\cmidrule{3-5}\cmidrule{7-9}\cmidrule{12-14}\cmidrule{16-18}\cmidrule{21-23}\cmidrule{25-27}
& Image 
& (a) & (b) & (c) && (a) & (b) & (c)  & \multirow{-1}{*}{\cellcolor{Goldenrod!20}Mean}
&
& (a) & (b) & (c) && (a) & (b) & (c) & \multirow{-1}{*}{\cellcolor{Goldenrod!20}Mean}
&
& (a) & (b) & (c) && (a) & (b) & (c) & \multirow{-1}{*}{\cellcolor{Goldenrod!20}Mean} \\ 
\midrule
\multirow{2}{*}{\rotatebox{90}{100}}


& LoRS~\cite{xu2024lors} 
& 19.4$^{\ast}$ & 10.0 & 9.2 && 15.6 & 8.7 & 8.2 & \cellcolor{Goldenrod!20}10.3
&& 27.4$^{\ast}$ & 6.5&7.1&&24.0&5.8&6.1&  \cellcolor{Goldenrod!20}9.9
&& 9.4$^{\ast}$ & 1.8&1.6&&6.8&1.2&1.2&\cellcolor{Goldenrod!20}2.5\\
    
& \multicolumn{1}{c}{\cellcolor{Goldenrod!20}Ours}
& 21.9$^{\ast}$ & 13.6 & 15.3 && 17.3 & 11.0 & 12.4 & \cellcolor{Goldenrod!20}\textbf{13.9}
&& 26.4$^{\ast}$&13.7&18.9&&20.8&11.6&15.3& \cellcolor{Goldenrod!20}\textbf{16.1}
&& 10.3$^{\ast}$&6.4&7.2&&7.0&4.6&5.2&\cellcolor{Goldenrod!20}\textbf{6.1}\\

\midrule

\multirow{2}{*}{\rotatebox{90}{200}}


& LoRS~\cite{xu2024lors} 
& 22.1$^{\ast}$ & 11.7 & 10.8 && 18.3 & 8.5 & 8.8 & \cellcolor{Goldenrod!20}11.6
&& 29.5$^{\ast}$&10.0&10.9&&22.7&8.3&8.9 & \cellcolor{Goldenrod!20}12.2
&& 11.4$^{\ast}$&1.6&3.2&&7.7&1.0&2.1&\cellcolor{Goldenrod!20}3.1\\

& \multicolumn{1}{c}{\cellcolor{Goldenrod!20}Ours}
& 24.3$^{\ast}$ & 14.7 & 18.9 && 19.4 & 10.8 & 14.2 & \cellcolor{Goldenrod!20}\textbf{15.6}
&& 28.2$^{\ast}$&15.0&20.8&&23.3&11.7&16.3& \cellcolor{Goldenrod!20}\textbf{17.4}
&& 13.1$^{\ast}$&9.2&10.2&&9.9&6.9&7.7&\cellcolor{Goldenrod!20}\textbf{8.7}\\

\midrule

\multirow{2}{*}{\rotatebox{90}{500}}


& LoRS~\cite{xu2024lors} 
& 25.0$^{\ast}$ & 9.9 & 9.5 && 19.3 & 6.3 & 6.2 & \cellcolor{Goldenrod!20}10.2
&& 31.6$^{\ast}$&15.3&13.4&&22.1&10.8&10.7& \cellcolor{Goldenrod!20}14.5
&& 13.5$^{\ast}$&1.4&1.3&&7.5&0.8&0.9&\cellcolor{Goldenrod!20}2.4\\

& \multicolumn{1}{c}{\cellcolor{Goldenrod!20}Ours}
& 26.2$^{\ast}$ & 13.8 & 20.5 && 20.5 & 10.0 & 16.2 & \cellcolor{Goldenrod!20}\textbf{16.2}
&& 30.6$^{\ast}$&17.4&23.3&&23.9&13.5&18.5& \cellcolor{Goldenrod!20}\textbf{19.3}
&& 15.3$^{\ast}$ & 9.8 & 11.3 && 11.4 & 7.6 & 8.9 & \cellcolor{Goldenrod!20} \textbf{9.8}\\

\bottomrule
\end{tabular}
}
\vspace{-8pt}
\end{table*}

\noindent\textbf{Qualitative Analysis.}
Fig.~\ref{fig:qualitative} presents a qualitative comparison of initial samples and our distilled data under the Flickr8k 100 pair setting. Since our method distills 768-dimensional text embeddings rather than raw captions, we visualize the text by retrieving the dataset caption whose embedding is closest to the distilled text vector. 
Although the distilled images visually appear to resemble the initial images, we observe an interesting pattern of noise-like high-frequency patches found in Fig.~\ref{fig:qualitative}, also observed from prior works~\cite{wu2024vldistill, xu2024lors}. We posit that this artifact difference in the image, coupled with the text embedding changes, jointly demonstrate the distillation effect from condensing large-scale real data into a smaller set.

\begin{figure}[t]
    \centering
    \begin{minipage}{\linewidth}
        \centering
        \captionof{table}{Compute statistics for different \# of data pairs.}
        \label{tab:distillation-efficiency}
        \scriptsize
        \setlength{\tabcolsep}{3pt}
        \renewcommand{\arraystretch}{1}
        \renewcommand{\aboverulesep}{1pt}
        \renewcommand{\belowrulesep}{1pt}
        \vspace{-3mm}
        \begin{tabular}{lcccc}
            \toprule
            \texttt{Flickr8k} &  & \multicolumn{3}{c}{Distillation} \\
            \cmidrule{3-5}
            Method & \# Pairs & Time/iter (sec) & \# iter till best & Total (min) \\          \midrule
            LoRS~\cite{xu2024lors} & & 5.43 & 850  & 76.93 \\
            \rowcolor{Goldenrod!20} Ours &    \multirow{-2}{*}{100}                 & 1.72 & 200  & 5.73 (\textcolor{ForestGreen}{$\downarrow$ \texttt{93\%}}) \\ 
            \midrule
            LoRS~\cite{xu2024lors} &  & 6.60 & 1,250 & 137.50 \\
            \rowcolor{Goldenrod!20} Ours &   \multirow{-2}{*}{200}                                   & 2.39 & 200  & 7.97 (\textcolor{ForestGreen}{$\downarrow$ \texttt{94\%}}) \\ 
            \midrule
            LoRS~\cite{xu2024lors} &  & 5.27 & 2,350 & 206.41 \\
            \rowcolor{Goldenrod!20} Ours &   \multirow{-2}{*}{500}                    & 4.41 & 50   & 3.68 (\textcolor{ForestGreen}{$\downarrow$ \texttt{98\%}})\\ 
            \bottomrule
        \end{tabular}
    \end{minipage}
    \vspace{0.4em}
    \begin{minipage}{\linewidth}
        \centering
        \includegraphics[width=.95\linewidth]{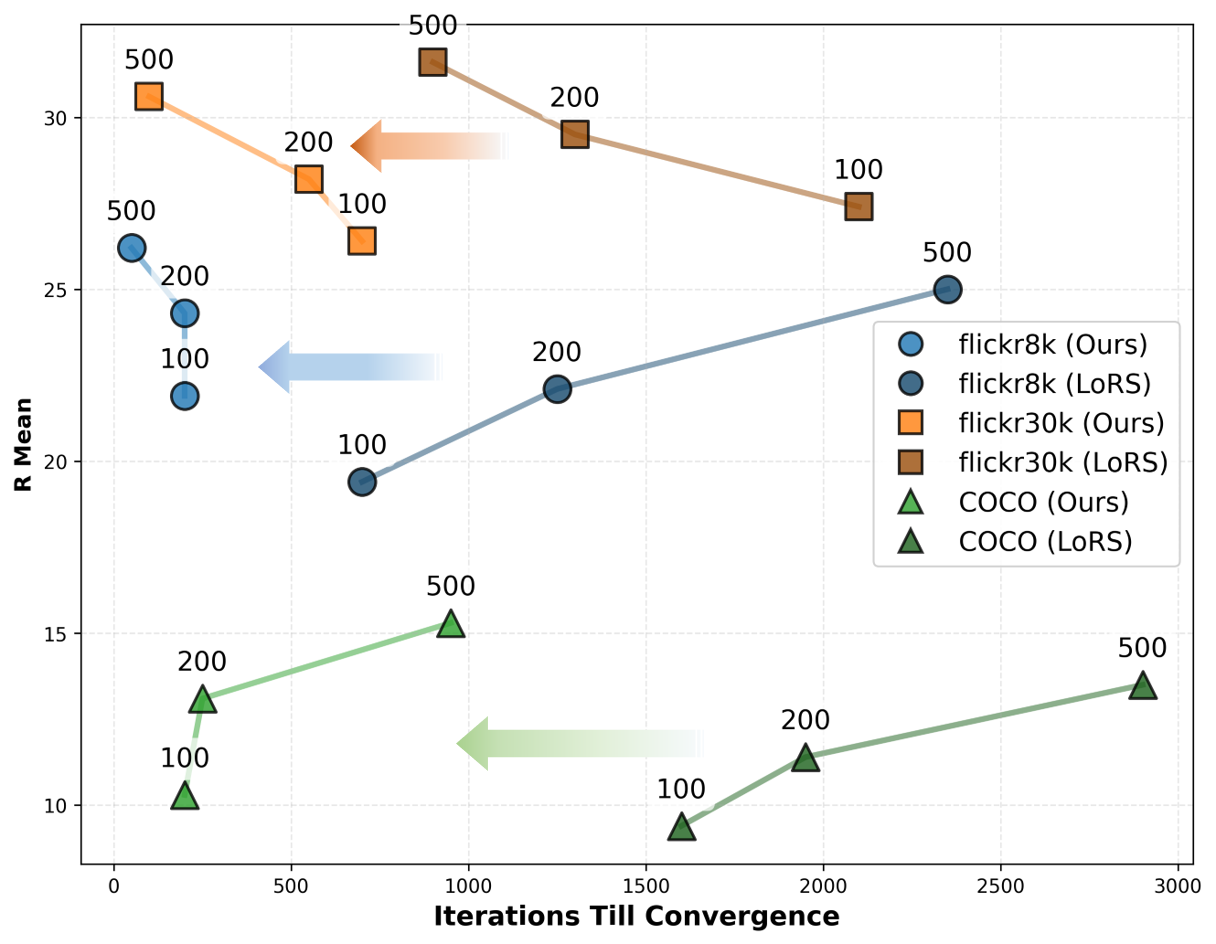}
        \vspace{-4mm}
        \captionof{figure}{Performance curve across datasets and data pairs. Ours consistently achieves higher performance at remarkably smaller iterations than the baseline~\cite{xu2024lors}.}
        \label{fig:performance_curve}
    \end{minipage}
    \vspace{-17pt}
\end{figure}

\subsection{Cross-Architecture Generalization}

To assess the generalizability of our MDM framework, we performed cross-architecture experiments in all three datasets under all budgets, and report them in Table~\ref{tab:cross}. All synthetic datasets are distilled using the NFNet+BERT configuration. We then train five different image–text encoder combinations using the distilled data and evaluate them on real test sets, reporting the averaged retrieval performance over IR/TR@K={1,5,10}. The "Mean" column corresponds to the average performance across the five cross-architecture settings. 

Across all datasets and budget settings, our method consistently achieves higher mean performance than LoRS, indicating stronger robustness to architecture changes. This improvement stems from the distinction between the two distillation paradigms: trajectory-matching methods, such as LoRS, inherit architecture-dependent biases from the source model's optimization path, limiting their transferability. In contrast, our distribution-matching formulation distills the underlying dataset-level multimodal distribution itself, producing synthetic samples that are less tied to a specific architecture and therefore generalize more reliably across different encoder combinations.

\subsection{Compute Efficiency}

We compare the computational cost of our method with LoRS during the distillation process in Table~\ref{tab:distillation-efficiency} and Fig.~\ref{fig:performance_curve}. 
The LoRS requires approximately 5.4 sec/iteration, whereas our method requires only 1.7 sec for each distillation update including per-iteration model initialization, nearly a 3$\times$ improvement in na\"ive per-iteration efficiency. 
This gap arises from the fundamental difference in computational structure. The MTT-based baseline repeatedly trains the student model to generate trajectories that match those of the real, and also performs \textit{bi-level} optimization to update the synthetic data, resulting in substantial overhead. 
In contrast, our MDM framework performs a \textit{single-level} optimization that directly updates the synthetic samples without trajectory reconstruction, making it considerably more lightweight.

Furthermore, as shown in Fig.~\ref{fig:performance_curve}, our method attains strong retrieval performance with dramatically fewer optimization iterations than LoRS; for example, in the Flickr8k–500 pair setting, convergence is achieved within only 50 iterations. 
As a result, the overall distillation process under MDM incurs dramatically lower computational cost than LoRS, highlighting the strong efficiency advantage of our method.

\subsection{Ablation Study and Further Discussion}

\noindent\textbf{Synthetic Data Initialization.}
We analyze the impact of synthetic data initialization in Table~\ref{tab:ablations}(a). 
Including this experiment, all ablation studies are conducted under the Flickr8k 100-pair setting.
In contrast to observations in classification-oriented dataset distillation~\cite{wang2018datasetdistillation, sajedi2023datadam, yin2023cda, cui2022dcbench}, initializing synthetic samples from Gaussian noise fails to produce meaningful retrieval performance, likely due to the higher complexity of Flickr8k/30k and COCO and the fine-grained alignment required by image–text retrieval.

Compared to random real-sample initialization, k-means–based initialization offers a clear advantage by selecting representative samples that better capture the dataset’s global structure. Moreover, clustering with joint features achieves the best results, insinuating the advantage of initialization reflecting the joint image-text structures more faithfully approximates the underlying multimodal distribution. In particular, even random sample initialization surpasses LoRS in the Flickr8k–100 pair setting, highlighting the effectiveness and robustness of our MDM framework.

\noindent\textbf{Model Initialization.}
Table~\ref{tab:ablations}(b) analyzes how the initialization of the frozen model—used to encode real and synthetic samples during distillation—affects performance. Initializing from the original pre-trained checkpoint yields performance comparable to random-sample initialization, indicating that the synthetic data cannot effectively capture the target-domain structure. In contrast, initializing from a model fine-tuned on the target real dataset yields clear improvements, highlighting the importance of distillation in an embedding space that reflects target-domain characteristics.


In addition, using at each distillation step a model formed by a random weighted sum of the pre-trained and fine-tuned checkpoints yields further improvements over using the fine-tuned model alone. This indicates that distillation benefits from both the final embedding space and the intermediate states between pre-trained and fine-tuned models, allowing synthetic data to leverage a wider range of representations. This observation is consistent with prior distribution-matching methods for classification~\cite{zhang2024dance, IDM}, which similarly utilize models from different training stages to better capture evolving training dynamics.

Our proposed strategy achieves the best performance: by interpolating across multiple fine-tuned experts in weight space, it avoids overfitting to any single representation configuration. It exposes synthetic data to a more diverse, balanced joint embedding space, suggesting that our MDM is most effective when the frozen model provides a broad yet target-relevant representation landscape.

\begin{table}[t]
\centering
\caption{Ablations across synthetic data and model initializations.}
\label{tab:ablations}
\vspace{-3mm}
  \begin{minipage}[t]{\linewidth}\centering
    \subcaption{Synthetic Data Initialization}
    \label{tab:ablation:syn_data_init}
    \setlength{\tabcolsep}{3pt}
    \renewcommand{\arraystretch}{1}
    \renewcommand{\aboverulesep}{1pt}
    \renewcommand{\belowrulesep}{1pt}
    \resizebox{0.85\textwidth}{!}{
    \begin{tabular*}{\linewidth}{@{\extracolsep{\fill}}lccccc@{}}
      \toprule
      &&& \multicolumn{3}{c}{K-means Clustering} \\\cmidrule{4-6}
      \multirow{1}{*}{Syn. Data Init.} & \multirow{1}{*}{Noise} & \multirow{1}{*}{Random} & Image & Text & \cellcolor{Goldenrod!20} Ours \\
      \midrule

      IR   & 0.6 & 18.6 & 18.9 & 18.6 & \cellcolor{Goldenrod!20}\textbf{19.7} \\
      TR   & 0.5 & 22.6 & 22.7 & 22.8 & \cellcolor{Goldenrod!20}\textbf{24.2} \\
      Mean & 0.5 & 20.6 & 20.8 & 20.7 & \cellcolor{Goldenrod!20}\textbf{21.9} \\
      \bottomrule
      \end{tabular*}}
  \end{minipage}
  
  \vspace{0.3em}
  
  \begin{minipage}[t]{\linewidth}\centering
    \subcaption{Model Initialization}
    \label{tab:ablation:model_init}
    \setlength{\tabcolsep}{3pt}
    \renewcommand{\arraystretch}{1}
    \renewcommand{\aboverulesep}{1pt}
    \renewcommand{\belowrulesep}{1pt}
    \resizebox{0.85\linewidth}{!}{
    \begin{tabular*}{\linewidth}{@{\extracolsep{\fill}}lcccc}
      \toprule
      Model Init. &      
      Pre-trained &
      Fine-tuned &
      Weighted Sum &
      \cellcolor{Goldenrod!20} Ours \\
      \midrule
      IR   & 5.1 & 14.3 & 17.6 & \cellcolor{Goldenrod!20}\textbf{19.7} \\
      TR   & 8.4 & 21.3 & 21.7 & \cellcolor{Goldenrod!20}\textbf{24.2} \\
      Mean & 6.8 & 17.8 & 19.6 & \cellcolor{Goldenrod!20}\textbf{21.9} \\
      \bottomrule
    \end{tabular*}}
  \end{minipage}
  \vspace{-10pt}
\end{table}

\setcounter{table}{4}
\begin{table}[t]
    \centering
    \caption{Ablation study on the loss components. We report the average retrieval scores over K=\{1,5,10\}.}
    \label{tab:ablation:loss}
    \vspace{-3mm}
    \setlength{\tabcolsep}{3pt}
    \renewcommand{\arraystretch}{1}
    \renewcommand{\aboverulesep}{1pt}
    \renewcommand{\belowrulesep}{1pt}
    \resizebox{.8\linewidth}{!}{
    \begin{tabular}{cccc|ccc}
    \toprule
         No. &  $\mathcal{L}_{\textrm{InfoNCE}}$ & $\mathcal{L}_{\textrm{agr}}$ & $\mathcal{L}_{\textrm{dis}}$ & IR & TR & Mean\\
     \midrule
        1 & \checkmark & \textcolor{gray!20}{\xmark} & \textcolor{gray!20}{\xmark} & 18.81 & 23.15 & 20.98 \\
        2 &  \checkmark & \checkmark & \textcolor{gray!20}{\xmark} & 18.82 &	23.23 &	21.02\\
        3 &  \checkmark & \textcolor{gray!20}{\xmark} & \checkmark & 19.22 & 23.84 & 21.53\\
        \rowcolor{Goldenrod!20} 
        4 &  \checkmark & \checkmark & \checkmark & \textbf{19.73} &	\textbf{24.15}	& \textbf{21.94} \\
        
    \bottomrule
    \end{tabular}}
    \vspace{-12pt}
\end{table}

\noindent\textbf{Component Analysis.}
Table~\ref{tab:ablation:loss} summarizes the ablation study on the loss components of MDM. The alignment loss $\mathcal{L}_{\text{InfoNCE}}$ provides the basic pairing consistency between synthetic images and texts. Adding the agreement loss $\mathcal{L}_{\text{agr}}$ encourages the image–text agreement of synthetic pairs to match that of real data, thereby improving cross-modal structural consistency slightly. On the other hand, adding the discrepancy loss solely boosts even further, implying that modeling the cross-modal gap bears more significance than the shared component.
Together, each complements by matching the image–text discrepancy between real and synthetic data, allowing the synthetic distribution to better capture the modality-specific separation present in real samples. Going beyond simple combination of data and model initializations, we remark that our MDM loss altogether act in synergy to achieve the state-of-the-art image-text retrieval performance.

\section{Conclusion}
\label{sec:conclusion}

In this paper, we present an extension of distribution matching to multimodal settings. Our proposed MDM is a lightweight, generalizable algorithm that effectively exploits the cross-modal alignment between vision and language features by operating directly in the joint embedding space and by leveraging geometry-aware objectives tailored to multi-modal representations.
Through this formulation, MDM avoids the heavy bi-level optimization and architecture-dependent biases, leading to strong improvements on cross-modal retrieval tasks across multiple datasets and substantial gains in cross-architecture generalization.

\vspace{-10pt}

\paragraph{Limitations and LLM Usage.}
We assume access to pretrained image and text encoders. While MDM leverages distribution matching to enhance cross-architecture generalization, it remains limited by this reliance on pretrained encoders.
We note that large language models (LLMs) have been used solely to assist with improving editing (\eg, grammar, spelling), formatting, and styling, without intervening in the development of original ideas.

\section*{Acknowledgments}
This work was supported by the National Research Foundation of Korea(NRF) grant funded by the Korea government(MSIT) (RS-2026-25473963) and the Institute of Information \& communications Technology Planning \& Evaluation (IITP) grant funded by the Korea government(MSIT) (No. RS-2024-00457882, AI Research Hub Project).

{
    \small
    \bibliographystyle{ieeenat_fullname}
    \bibliography{main}
} 

\clearpage
\appendix
\setcounter{page}{1}
\maketitlesupplementary

\setcounter{page}{1}
\setcounter{figure}{0}
\setcounter{table}{0}
\setcounter{equation}{0}
\renewcommand{\thefigure}{S\arabic{figure}}
\renewcommand{\thetable}{S\arabic{table}}
\renewcommand{\theequation}{S\arabic{equation}}
\renewcommand{\thesection}{S\arabic{section}}

\noindent In this supplementary material, we elaborate on the details of our method and experiments and provide additional results with further analyses. We remark that throughout the manuscript, we intend to work with features denoted as $z^v$ for the image and $z^t$ for the text, the angle between the two displacement vectors as $_\angle \phi (\cdot,\cdot)$, and the distance (discrepancy) function as $\phi(\cdot,\cdot)$.
\noindent The list of contents is as follows:
\vspace{1mm}
\begin{enumerate} \itemsep 0.3em
    \item \textbf{Generalizing Unimodal DM to Multimodal} (Sec.~\ref{sec:supp:mdm_derivation})
    \item \textbf{Distinctions of Our Method} (Sec.~\ref{sec:supp:distinctions})
    \item \textbf{Experimental Details} (Sec.~\ref{sec:supp:exp_details}) \\
    - Selection of datasets of various scales (Sec.~\ref{sec:supp:dataset_selection}) \\
    - Reasons for underperformance on Flickr30k (Sec.~\ref{sec:supp:underperformance}) \\
    - Limitations of the baseline (Sec.~\ref{sec:supp:limitations_baseline}) \\
    - Further analysis on initialization (Sec.~\ref{sec:supp:analysis}) \\
    - Implementation details (Sec.~\ref{sec:supp:implementation}) \\
    - Sensitivity analysis (Sec.~\ref{sec:supp:sensitivity}) \\
    - Full retrieval results over multiple runs (Sec.~\ref{sec:supp:full_results}) \\
    - Full algorithm (Sec.~\ref{sec:supp:algorithm}) \\
\end{enumerate}

\section{MDM Formulation}
\label{sec:supp:mdm_derivation}

\paragraph{Generalizing Unimodal DM to Multimodal.}\quad
In our approach, we consider a multimodal dataset $\mathcal{D}_{\mathrm{real}} = \{(x_i,t_i)\}_{i=1}^{B}$ of image-text pairs and a much smaller synthetic dataset $\mathcal{D}_{\mathrm{syn}} = \{(\tilde x_j,\tilde t_j)\}_{j=1}^{\tilde B}$ with $|\mathcal{D}_{\mathrm{syn}}| \ll |\mathcal{D}_{\mathrm{real}}|$. A unified image-text model $\Psi(\cdot,\cdot)$ maps an image-text pair $(x,t)$ into a joint feature space and is composed of a pretrained image encoder $\theta^{v}$, and a pretrained text encoder $\theta^{t}$ with a projection layer.

Our goal is to acquire an optimal set of distilled set $\mathcal{D}_{\mathrm{syn}}^{\star}$ via the multimodal distribution matching (MDM) objective (Eq. \textcolor{cvprblue}{2}):
\vspace{-1mm}
\begin{equation}
\resizebox{\linewidth}{!}{$
\mathcal{D}_{\mathrm{syn}}^{\star} =
\displaystyle\operatorname*{arg\,min}_{\mathcal{D}_{\mathrm{syn}}}
\phi\Big(
\underbrace{
\mathbb{E}_{(X,T)\sim\mathcal{D}_{\mathrm{real}}}
\big[\Psi(X,T)\big]
}_{\text{real joint}}
,
\underbrace{
\mathbb{E}_{(\tilde{X},\tilde{T})\sim\mathcal{D}_{\mathrm{syn}}}
\big[\Psi(\tilde{X},\tilde{T})\big]
}_{\text{synthetic joint}}
\Big),
$}\nonumber
\vspace{-1mm}
\end{equation}
\noindent where $\phi: \mathbb{R}^d \times \mathbb{R}^d \to [0,\infty)$ is a nonnegative discrepancy function measuring the distance between two mean joint 
feature vectors.
Below, we justify this objective as a natural multimodal extension of the standard unimodal distribution matching formulation.


We first interpret the real and synthetic datasets as empirical distributions to ease the understanding of their distributions. The real data $\mathcal{D}_{\mathrm{real}}$ induces the empirical distribution:
\vspace{-3mm}
\begin{equation}
\widehat P_{XT}^{\mathrm{real}}
=
\frac{1}{B}\sum_{i=1}^{B} \delta_{(x_i,t_i)},
\end{equation}
\noindent where $\delta_{(x_i,t_i)}$ is the Dirac measure at the pair $(x_i,t_i)$.
Similarly, the synthetic dataset $\mathcal{D}_{\mathrm{syn}}$ induces
\vspace{-2mm}
\begin{equation}    
\widehat Q_{XT}^{\mathrm{syn}}
=
\frac{1}{\tilde B}\sum_{j=1}^{\tilde B} \delta_{(\tilde x_j,\tilde t_j)}.
\end{equation}
\noindent For any fixed joint feature extractor $\Psi$, the expectations of $\Psi$ under these empirical distributions are simply empirical means:
\vspace{-2mm}
\begin{align}
\mathbb{E}_{(X,T)\sim\widehat P_{XT}^{\mathrm{real}}}
\big[\Psi(X,T)\big]
&=
\frac{1}{B}\sum_{i=1}^{B} \Psi(x_i,t_i),
\\
\mathbb{E}_{(\tilde X,\tilde T)\sim\widehat Q_{XT}^{\mathrm{syn}}}
\big[\Psi(\tilde X,\tilde T)\big]
&=
\frac{1}{\tilde B}\sum_{j=1}^{\tilde B} \Psi(\tilde x_j,\tilde t_j).
\end{align}
Thus, the expectations appearing in Eq.\textcolor{cvprblue}{2} can be interpreted as dataset-wise averages of joint features produced by $\Psi$ over the real and synthetic datasets, respectively.
\noindent We can then define the real and synthetic mean joint features as
\begin{align}
\mu_{\mathrm{real}} &:= \mathbb{E}_{(X,T)\sim P_{XT}^{\mathrm{real}}} \big[\Psi(X,T)\big], \\
\mu_{\mathrm{syn}}(\mathcal{D}_{\mathrm{syn}}) &:= \mathbb{E}_{(\tilde X, \tilde T)\sim Q_{XT}^{\mathrm{syn}}} \big[\Psi(\tilde X,\tilde T)\big],
\end{align}
\noindent where we denote the true data distribution over image-text pairs $(X, T)$ as $P_{XT}^{\mathrm{real}}$, and the distribution represented by $\mathcal{D}_{\mathrm{syn}}$ as $Q_{XT}^{\mathrm{syn}}$.
The goal of multimodal DM is then to find a synthetic dataset $\mathcal{D}_{syn}$ whose induced distribution $Q_{XT}^{\mathrm{syn}}$ yields a mean feature vector $\mu_{\mathrm{syn}}(\mathcal{D}_{\mathrm{syn}})$ that is as close to the real mean $\mu_{\mathrm{real}}$ under some discrepancy function $\phi(\cdot,\cdot)$:
\begin{equation}
\mathcal{D}_{\mathrm{syn}}^{\star}
\in
\operatorname*{arg\,min}_{\mathcal{D}_{\mathrm{syn}}}
\phi\big(
\mu_{\mathrm{real}},
\mu_{\mathrm{syn}}(\mathcal{D}_{\mathrm{syn}})
\big).
\end{equation}
Replacing the population expectations by their empirical counterparts $\widehat P_{XT}^{\mathrm{real}}$ and $\widehat Q_{XT}^{\mathrm{syn}}$, we obtain
\begin{align}
\mathbb{E}_{(X,T)\sim\mathcal{D}_{\mathrm{real}}}[\Psi(X,T)]
\approx
\mu_{\mathrm{real}},
\\
\mathbb{E}_{(\tilde X,\tilde T)\sim\mathcal{D}_{\mathrm{syn}}}[\Psi(\tilde X,\tilde T)]
\approx
\mu_{\mathrm{syn}}(\mathcal{D}_{\mathrm{syn}}).
\end{align}
\noindent For fixed $\Psi$ and $\mathcal{D}_{\mathrm{real}}$, the real mean $\mu_{\mathrm{real}}$ is constant and serves as the target, while the synthetic mean $\mu_{\mathrm{syn}}(\mathcal{D}_{\mathrm{syn}})$ depends on the content of $\mathcal{D}_{\mathrm{syn}}$ and is optimized by updating the synthetic pairs.
We remark that this is equivalent to the unimodal DM formulation if we replace the joint image-text feature $\Psi(x,t)$ with a unimodal encoder $\theta_{0}(x)$ as: 
\begin{equation}
\mathcal{D}_{\mathrm{syn}}^{\star} =
\displaystyle\operatorname*{arg\,min}_{\mathcal{D}_{\mathrm{syn}}}
\phi\Big(
\mathbb{E}_{X\sim\mathcal{D}_{\mathrm{real}}}\big[\theta_{0}(X)\big]
,
\mathbb{E}_{\tilde X\sim\mathcal{D}_{\mathrm{syn}}}\big[\theta_{0}(\tilde X)\big]
\Big),
\end{equation}

\noindent for some distance function $\phi(\cdot,\cdot)$ for unimodal distributions.

Under these choices, the random variable $(X, T)$ effectively reduces to $X$, and the mean joint features become mean image features.
Hence, the proposed multimodal DM objective strictly generalizes the standard image-only DM objective by extending the feature space from $\theta_{0}(x)$ to the joint representation $\Psi(x,t)$ and by allowing a general discrepancy function.

\section{Distinctions of Our Method}\label{sec:supp:distinctions}
Our MDM method first seeds synthetic image--text pairs by running K-means clustering on the concatenated joint features $[z^v; z^t]$. It then optimizes these synthetic pairs so that the spherical distributions of an agreement vector $u$ and a discrepancy vector $g$ match those of the real data. The agreement $u$ captures shared image--text content and is obtained from a normalized combination of $z^v$ and $z^t$ on the unit hypersphere. The discrepancy $g$ encodes the modality gap between image and text and is obtained from a normalized difference of $z^v$ and $z^t$ on the same hypersphere. Matching the real and synthetic distributions in both $u$ and $g$ encourages the synthetic set to capture \textit{architecture-agnostic joint semantics} rather than reproducing individual training trajectories. Here, interestingly, we observe from the ablation study in Table~\textcolor{cvprblue}{5} that matching the distribution of $g$ improves retrieval performance by a larger gap than that from matching $u$. This indicates that MDM primarily encourages the learning of how captions \textit{deviate} from images through the global structure of the gap distribution, rather than focusing on \textit{shared} semantics. 

\subsection{Synthetic Data Initialization}
Coreset-based initializations such as K-center and herding select real image–text pairs that approximately cover the encoder feature 
space under max–min radius or greedy moment-matching criteria, and then reuse these as seeds for optimization under the same MDM 
objective. However, these heuristics neglect the structure of the image-text agreement and discrepancy, and encourage only marginal 
coverage, rather than explicitly targeting the joint semantic modes that are crucial for effective cross-modal retrieval. In contrast, 
our initialization performs K-means clustering directly in the \textit{joint} feature space and assigns the sample nearest the cluster centroid to each synthetic pair, anchoring the synthetic parameters near representative joint centroids. This joint-feature-aware seeding reduces the burden on MDM to relocate poorly placed seeds and instead focuses optimization on fine-grained refinement around already well-positioned prototypes, yielding synthetic datasets that more faithfully approximate the real joint distribution. Empirically, this leads to consistently higher image-text retrieval performance than coreset-seeded variants, shown in Table~\ref{tab:supp:coreset}.
\vspace{-2mm}
\begin{table}[!ht]
    \centering
    \caption{Performance comparison on Ours seeded with different synthetic data initialization strategies.}
    \vspace{-3mm}
    \label{tab:supp:coreset}
    \setlength{\tabcolsep}{10pt}
    \renewcommand{\arraystretch}{1}
    \renewcommand{\aboverulesep}{2pt}
    \renewcommand{\belowrulesep}{2pt}
    \resizebox{.99\linewidth}{!}{
    \begin{tabular}{lccccc}
    \toprule
        & \multirow{2}{*}{\textbf{K-center}} & \multirow{2}{*}{\textbf{Herding}} & \cellcolor{Goldenrod!20} \textbf{Joint-feature K-Means} \\
        & & & \cellcolor{Goldenrod!20} \textbf{Clustering (Ours)}  \\
    \midrule
         IR     & 19.5 &19.2 & \cellcolor{Goldenrod!20} \textbf{19.7}\\
         TR     & 22.1 &23.9 & \cellcolor{Goldenrod!20} \textbf{24.2}\\
         Mean   & 20.8 &21.5 & \cellcolor{Goldenrod!20} \textbf{21.9}\\
    \bottomrule
    \end{tabular}}
    \vspace{-8pt}
\end{table}

\subsection{Model Initialization}
Our method employs a dynamic multimodal weight-initialization scheme that merges a pretrained image encoder–text projector pair with $N$ finetuned counterparts randomly sampled from a pool of experts, each updated at every training iteration as with vision-only DM methods~\cite{DM,IDM,zhang2024dance}. Inspired by the angular, layer-wise interpolation strategy of Model Stock~\cite{jang2024modelstock} (originally proposed for robust \textit{unimodal} classification), we compute layer-wise interpolation coefficients from the angles between the displacement vectors of each finetuned expert anchored on the model composed of a pretrained image encoder and a randomly initialized text projector. Using these coefficients, we then merge them in the weight space, with the goal of shifting the pretrained anchor towards real data distribution, implicitly guided by these experts. To this, we add an additional global weighting factor, $\alpha$, to further modulate these layer-wise ratios, effectively controlling the trade-off between expert-specific bias and generic structure in a fine-grained manner. Unlike the static, one-shot merge in Model Stock that is used as the final predictor, the merged multimodal encoder in our framework serves as a stochastic, expert-pool–aware initialization that is refreshed at every distillation step and then optimized under the MDM objective. This construction extends \cite{jang2024modelstock} from unimodal to multimodal (image encoder and text projector) weight fusion and allows synthetic data to be optimized in a more diverse region of the joint image–text weight space, effectively stabilizing and improving the optimization process.


\subsection{Geodesic Kernel Formulation}
Geodesic distance on the sphere measures the shortest path along the sphere with respect to the manifold's curvature, hence yielding the \textit{intrinsic} distance. Our intuition to employ the geodesic distance stems from \cite{mei2025geomm}, which highlighted its advantage for understanding the complex geometric structure of multimodal data. Since typical InfoNCE loss and retrieval tasks are primarily driven by angular similarity (\eg, cosine), our \textit{geodesic} perspective using the geodesic kernel energy as formulated in Eq. \textcolor{cvprblue}{9} aligns well with the \textit{angular} distributions of multimodal data.

\noindent\textbf{Kernel variants.}
For two $d$-dim feature vectors $a,b \in \mathbb{R}^d$ and a bandwidth $\sigma > 0$, we consider the following radial basis function (RBF) kernels with different angular distance functions that induce the same topology on $\mathbb{S}^{d-1}$, where we demonstrate that the geodesic kernel performs the best in Table~\ref{tab:supp:kernel}. Although the image retrieval score improvements appear incremental, our geodesic kernel dramatically increases text retrieval scores.

\begin{itemize}

    \item \textbf{Laplacian RBF kernel.}
    We also consider a Laplacian RBF kernel based on the L1-distance in the ambient space:
    \begin{equation}
        k_{\mathrm{laplacian}}(a,b)
        \;=\;
        \exp\!\left(
            -\frac{\lVert a - b \rVert_1}{\sigma}
        \right).
    \end{equation}
    
    \item \textbf{Chordal RBF kernel on the unit hypersphere.}\quad
    Similar to the well-established Euclidean distance, the chordal distance measures the straight-line chord distance between the two unit-normalized points, but restricted to points on a sphere. To force the features onto the sphere, we first L2-normalize the features to lie on the unit hypersphere,
    $\hat{a} = a / \lVert a \rVert_2$, $\hat{b} = b / \lVert b \rVert_2$,
    and define the \emph{chordal distance} between $\hat{a}$ and $\hat{b}$ as
    \begin{equation}
        d_{\mathrm{chord}}(\hat{a}, \hat{b}) = \lVert \hat{a} - \hat{b} \rVert_2 = \sqrt{2 - 2\langle \hat{a}, \hat{b} \rangle} ,
    \end{equation}
    where $\langle \cdot,\cdot \rangle$ denotes inner product.
    The corresponding chordal RBF kernel is then:
    \begin{equation}
    \resizebox{.95\linewidth}{!}{$
        k_{\mathrm{chord}}(a,b) = \exp\!\left( -\frac{d_{\mathrm{chord}}(\hat{a}, \hat{b})^2}{2\sigma^2} \right) = \exp\!\left( -\frac{2 - 2\langle \hat{a}, \hat{b} \rangle}{2\sigma^2} \right).
    $}
    \end{equation}
    Intuitively, chordal distance measures the Euclidean length of the straight-line chord in the ambient space connecting two points on the sphere.
    \item \textbf{Geodesic RBF kernel on the unit hypersphere.}
    The intrinsic (geodesic) distance on the unit hypersphere is given by the angular distance along the arc on the surface of the sphere as:
    \begin{equation}
        d_{\mathrm{geo}}(\hat{a}, \hat{b})
        = \arccos\!\big( \langle \hat{a}, \hat{b} \rangle \big) \;\in [0,\pi].
    \end{equation}
    Using this intrinsic distance, we define the geodesic Gaussian kernel
    \begin{equation}
    \resizebox{.5\linewidth}{!}{$
        k_{\mathrm{geo}}(a,b) = \exp\!\left( -\frac{d_{\mathrm{geo}}(\hat{a}, \hat{b})^2}{2\sigma^2} \right).
    $}
    \end{equation}
 Although this kernel is not guaranteed to be positive definite on $\mathbb{S}^{d-1}$ in general, we use the resulting quantity as a \emph{geodesic kernel energy} to encourage the alignment of real and synthetic feature distributions in the unit hypersphere.
\end{itemize}

\vspace{-2mm}
\begin{table}[!t]
    \centering
    \caption{Performance by different types of RBF Kernel.}
    \vspace{-3mm}
    \label{tab:supp:kernel}
    \setlength{\tabcolsep}{15pt}
    \renewcommand{\arraystretch}{1}
    \renewcommand{\aboverulesep}{2pt}
    \renewcommand{\belowrulesep}{2pt}
    \resizebox{.99\linewidth}{!}{
    \begin{tabular}{lccccc}
    \toprule
         \textbf{$k(\cdot,\cdot)$} & \textbf{Laplacian} & \textbf{Chordal}  &  \cellcolor{Goldenrod!20} \textbf{Geodesic (Ours)}  \\
    \midrule
         IR     & 18.3 & 19.5 & \cellcolor{Goldenrod!20} \textbf{19.7}  \\
         TR     & 22.4 & 23.5 & \cellcolor{Goldenrod!20} \textbf{24.2} \\
         Mean   & 20.3 & 21.5 & \cellcolor{Goldenrod!20} \textbf{21.9} \\
    \bottomrule
    \end{tabular}}
    \vspace{-8pt}
\end{table}

\subsection{Cross-Modal Agreement and Discrepancy}
\begin{figure}[!t]
    \centering
    \includegraphics[width=\linewidth]{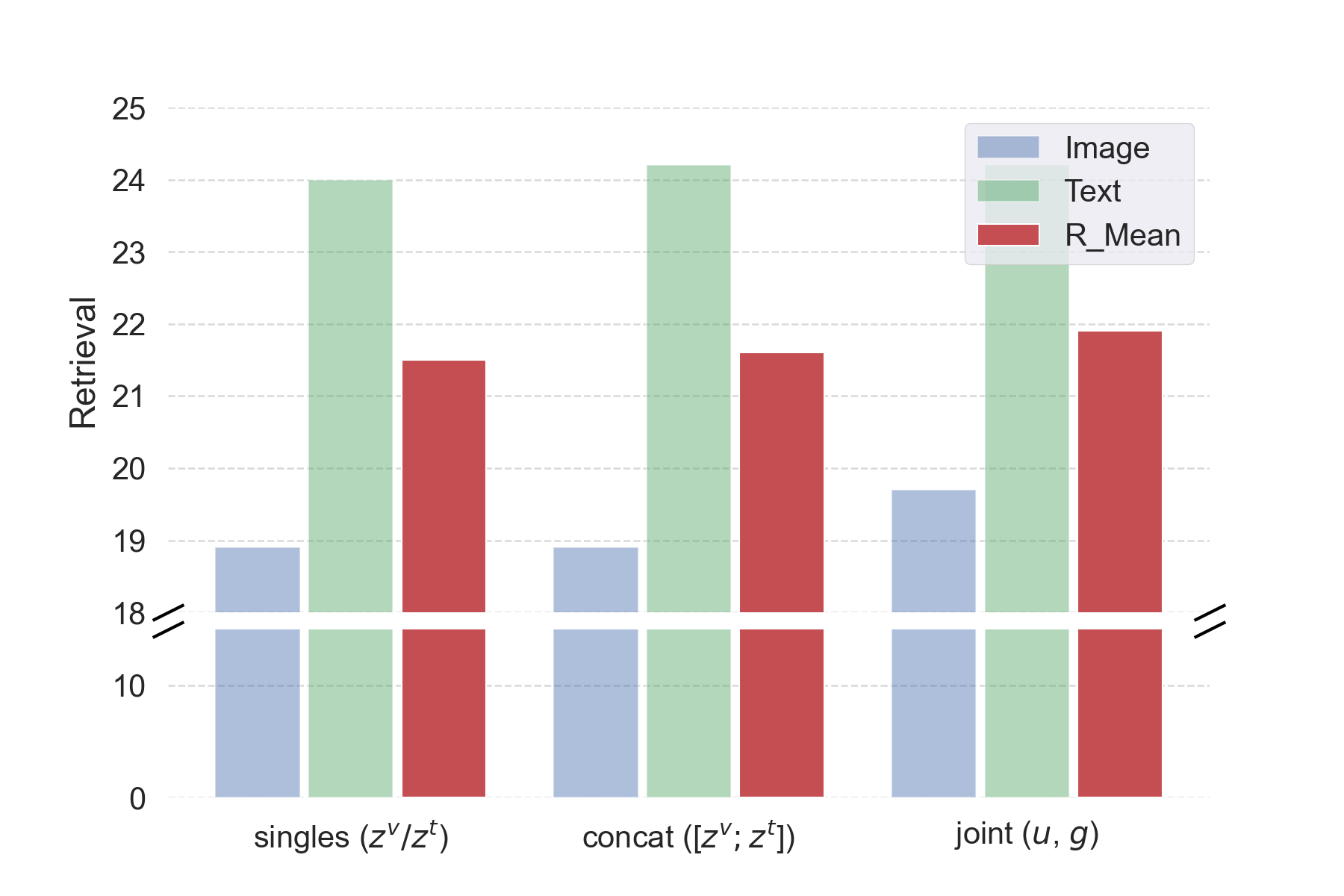}
    \vspace{-9mm}
    \caption{Comparison of matching geodesic kernel energy on (i) each image and text features separately, (ii) concatenated image-text features only, and (iii) our joint cross-modal agreement and discrepancy features.}
    \label{fig:supp:joint_features}
    \vspace{-8pt}
\end{figure}

To better capture the structure of multimodal representations, we construct the joint image-text features as a \emph{cross-modal agreement} vector $u$ and a \emph{cross-modal discrepancy} vector $g$. The agreement component encodes \textbf{modality-shared semantics} (objects, actions, coarse scene layout), whereas the discrepancy component models \textbf{modality-specific} information, \ie, the inherent ``modality gap''. This gap is non-negligible in image-text data: a single image can be paired with multiple, diverse captions (as in Flickr8k, Flickr30k, MS-COCO datasets), and the same visual concept can be expressed in many textual forms, leading to a structured one-to-many relationship between the two modalities. As reported in Fig.~\ref{fig:supp:joint_features}, reducing geodesic kernel energy on each modality independently (``\textit{singles}'') or on na\"vely concatenated joint features (``\textit{concat}'') fails to \textit{fully} account for this structure. In contrast, our joint optimization over $u$ and $g$ (``\textit{joint}'') explicitly matches both modality-shared semantics and modality-specific gaps, yielding better preservation of joint image-text relationships and consistently higher retrieval performance.

\section{Experimental Details} \label{sec:supp:exp_details}

\subsection{Deliberate Selection of Datasets}\label{sec:supp:dataset_selection}
\begin{figure}[!t]
\centering
\includegraphics[width=.99\linewidth]{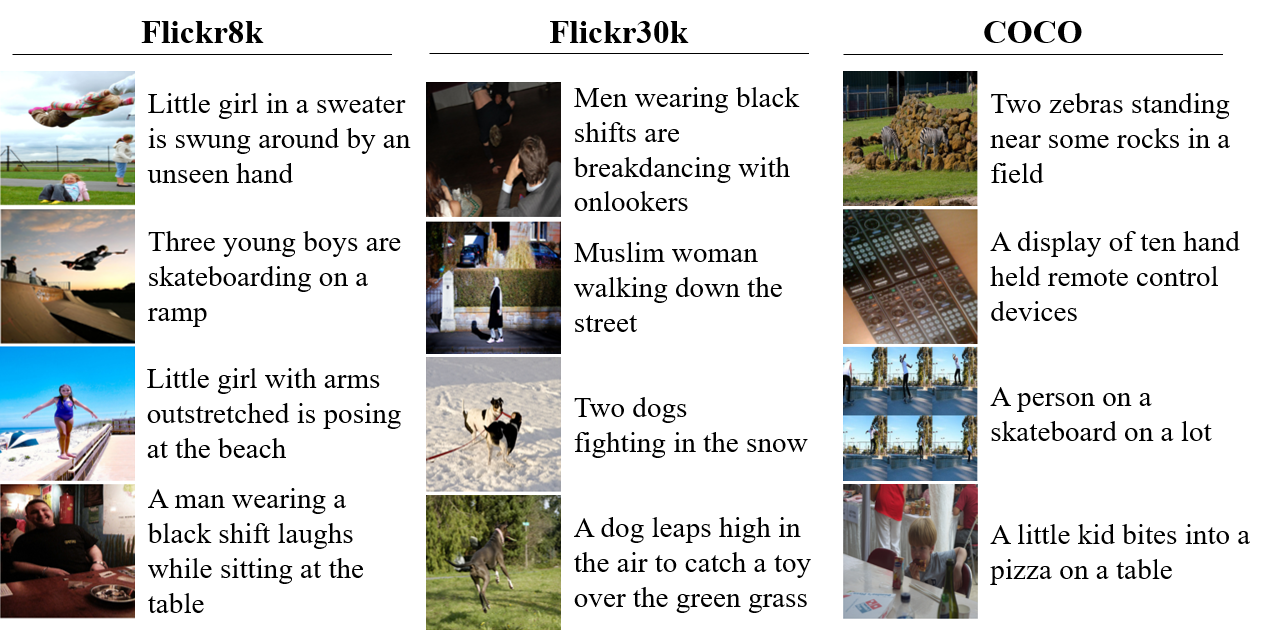}
\vspace{-3mm}
\caption{Examples of image-text datasets consisting of natural scene images and corresponding captions.}
\label{supp:fig:dataset_examples}
\vspace{-10pt}
\end{figure}

To study how our multimodal dataset distillation behaves as the training corpus grows, we deliberately choose three image–text benchmark datasets as exemplified in Fig.~\ref{supp:fig:dataset_examples} that particularly \textit{differ in scale}: Flickr8k~\cite{hodosh2013flickr8k} ($\approx$8k pairs), Flickr30k~\cite{young2014flickr} ($\approx$30k pairs) and MS-COCO~\cite{COCO} ($\approx$123k pairs). These datasets of increasing scale allow us to probe whether the same distillation procedure continues to yield meaningful compression and retrieval performance as we move from a small to a substantially larger dataset, while keeping the task and evaluation protocol comparable. Across this entire scale range, our method remains highly competitive, effectively condensing real data into compact synthetic subsets at substantially lower computational resource cost than existing baselines.

\subsection{Reasons for Underperformance on Flickr30k}\label{sec:supp:underperformance}
In Table~\textcolor{cvprblue}{1} of the main paper, we observe that on Flickr30k, a mid-scale dataset with relatively low redundancy, our text retrieval (TR) performance slightly underperforms the LoRS baseline. Each image in Flickr30k has multiple \textit{locally diverse} captions, and many images are visually similar while differing only in subtle relational or attribute-level details. Under MDM, the real gap vectors $\{g_i\}$ associated with such visually and semantically related image--caption pairs are pulled toward a small number of synthetic gap prototypes. This averaging of gap modes reduces the margin between the ground-truth captions and near-duplicate but different captions for a given image. Caption-level discrimination in text retrieval on Flickr30k, therefore, becomes harder than for LoRS~\cite{xu2024lors}, which preserves more instance-level detail via low-rank similarity and maintains sharper caption margins.

In contrast, the Flickr8k dataset is about 4$\times$ smaller and has a sparser caption space, with fewer near-duplicate captions per image and fewer confusing alternatives at evaluation time. In this low-data regime, the same smoothing of the gap distribution acts mainly as regularization. It suppresses unstable or idiosyncratic gaps instead of collapsing many truly distinct modes, which improves generalization relative to LoRS. On COCO, the largest and most redundant dataset in our evaluation, many caption patterns and gap structures repeat across a large number of images. In this setting, MDM compresses these repeated gap patterns into a compact set of prototypes and achieves more efficient coverage of typical multimodal relations than LoRS. This explains why our method outperforms the baseline on COCO despite operating at a stronger effective compression rate.

\subsection{Limitations of the Baseline~\cite{xu2024lors}} \label{sec:supp:limitations_baseline}
The LoRS baseline constructs synthetic data by enforcing low-rank similarity between the behavior of real and synthetic examples on a fixed source architecture. This ties the distilled set closely to the geometry and inductive biases of a particular architecture and favors instance-level reproduction of that model's gradients and feature updates. LoRS does not explicitly model the multimodal feature distribution. Instead, it approximates the real data distributions through a low-rank subspace of parameter updates. As a result, LoRS can preserve fine caption-level distinctions that benefit same-architecture text retrieval on Flickr30k. Yet, the synthetic data often remains specialized to the source model's decision boundaries and transfers poorly to different encoders. In contrast, our MDM approach operates directly in a joint feature space and matches distributions over both agreement and discrepancy between real and synthetic pairs. This induces the synthetic set to capture architecture-agnostic structure in the joint image-text manifold rather than reproducing low-rank gradient behavior tailored to a single architecture. The resulting synthetic data provide a more faithful distribution-level approximation of multimodal semantics and modality gaps, which accounts for the superior cross-architecture generalization, even though LoRS retains a small advantage in same-architecture text retrieval on Flickr30k.

\begin{figure*}[!h]
    \centering
    \includegraphics[width=.99\linewidth]{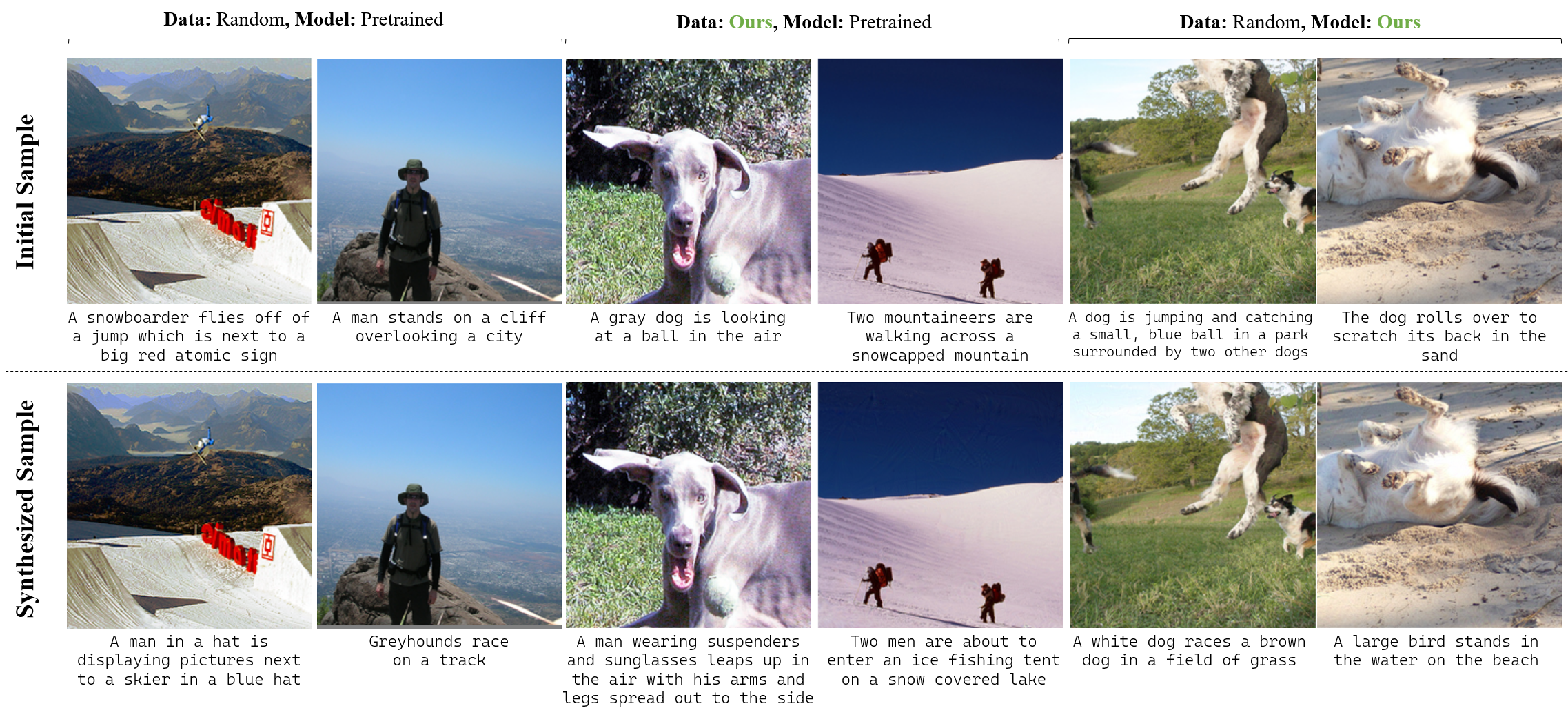}
    \vspace{-3mm}
    \caption{Qualitative comparisons for the ablation studies with different data and model initializations.}
    \label{fig:supp_ablation} 
    \vspace{-8pt}
\end{figure*}
\subsection{Further Analysis on Initialization} \label{sec:supp:analysis}
In Fig.~\ref{fig:supp_ablation}, we further inspect how different choices of data and model initialization affect the distilled (synthesized) samples. Across all three settings, the synthesized captions change locally relative to their initial counterparts, making the semantic effect of distillation more apparent on the text side. With random data and a pretrained model (\textit{left}), the generated captions are often clearly misleading. For instance, they hallucinate or misidentify people and scene context, indicating poor image–text alignment. When we keep the model pretrained but initialize the synthetic set using our clustering-based joint-feature seeding (\textit{middle}), the captions become slightly more faithful: core objects such as people are better captured, yet high-level scene understanding (\eg, “mountain” vs. “lake”) is still inaccurate. In contrast, using randomly sampled real data together with our mixed model initialization (\textit{right}) yields captions that more reliably track the local objects and actions in the image, although some residual mismatches remain. These qualitative trends align with the quantitative ablation studies reported in Table \textcolor{cvprblue}{4} of the main paper, where combining both our data initialization and model initialization gives the strongest overall image–text matching performance. We refer to Fig. \textcolor{cvprblue}{3} of our main paper for the qualitative results of Ours altogether.

\subsection{Implementation Details} \label{sec:supp:implementation}

Unless otherwise noted, we conducted all ablation studies on the Flickr8k~\cite{hodosh2013flickr8k} dataset using 100 pairs for uniformity. 

\subsubsection{System Configuration}
We carried out all the experiments on a Linux Machine equipped with an Intel Xeon(R) Silver 4210R CPU and a single NVIDIA RTX A6000 GPU, following the hyperparameter settings listed in Table~\ref{tab:supp:hyperparameters}. The software stack includes Python 3.10, PyTorch 2.6.0, and Torchvision 0.21.0, with support for CUDA 11.8 and cuDNN 9.1.0.

\begin{table}[!ht]
    \centering
    \caption{Hyperparameters for different experiments.}
    \label{tab:supp:hyperparameters}    
    \vspace{-3mm}
    \setlength{\tabcolsep}{3pt}
    \renewcommand{\arraystretch}{1}
    \renewcommand{\aboverulesep}{1pt}
    \renewcommand{\belowrulesep}{1pt}
    \resizebox{\linewidth}{!}{
    \begin{tabular}{l ccc ccc ccc}
    \toprule
    Hyperparam. & \multicolumn{3}{c}{Flickr8k~\cite{hodosh2013flickr8k}} &  \multicolumn{3}{c}{Flickr30k~\cite{young2014flickr}} &  \multicolumn{3}{c}{COCO~\cite{COCO}} \\
    \midrule
         \# Pairs & 100 & 200 & 500 & 100 & 200 & 500 & 100 & 200 & 500  \\
         Batch size & 64 & 64 & 64 & 64 & 64 & 64 & 64 & 64 & 64 \\
         LR$_{\mathrm{img}}$ & 100 & 100 & 1000 & 100 & 1000 & 1000 & 1000 & 1000 & 5000 \\ 
         LR$_{\mathrm{txt}}$ & 100 & 100 & 1000 & 100 & 1000 & 1000 & 1000 & 1000 & 5000 \\ 
         $\lambda_{\textrm{agr}}$ &  \multicolumn{9}{c}{\textemdash\textemdash\textemdash\textemdash\textemdash\textemdash\textemdash\textemdash\qquad
         0.8
         \qquad\textemdash\textemdash\textemdash\textemdash\textemdash\textemdash\textemdash\textemdash} \\
         $\sigma_{\textrm{agr}}$ & \multicolumn{9}{c}{\textemdash\textemdash\textemdash\textemdash\textemdash\textemdash\textemdash\textemdash\qquad
         0.5
         \qquad\textemdash\textemdash\textemdash\textemdash\textemdash\textemdash\textemdash\textemdash} \\
         $\lambda_{\textrm{dis}}$ & \multicolumn{9}{c}{\textemdash\textemdash\textemdash\textemdash\textemdash\textemdash\textemdash\textemdash\qquad
         0.8
         \qquad\textemdash\textemdash\textemdash\textemdash\textemdash\textemdash\textemdash\textemdash} \\
         $\sigma_{\textrm{dis}}$ & \multicolumn{9}{c}{\textemdash\textemdash\textemdash\textemdash\textemdash\textemdash\textemdash\textemdash\qquad
         0.5
         \qquad\textemdash\textemdash\textemdash\textemdash\textemdash\textemdash\textemdash\textemdash} \\
         $\alpha$ & \multicolumn{9}{c}{\textemdash\textemdash\textemdash\textemdash\textemdash\textemdash\textemdash\textemdash\qquad
         0.5
         \qquad\textemdash\textemdash\textemdash\textemdash\textemdash\textemdash\textemdash\textemdash} \\
         min expert epoch & \multicolumn{9}{c}{\textemdash\textemdash\textemdash\textemdash\textemdash\textemdash\textemdash\textemdash\qquad
         1
         \qquad\textemdash\textemdash\textemdash\textemdash\textemdash\textemdash\textemdash\textemdash} \\
         max expert epoch & \multicolumn{9}{c}{\textemdash\textemdash\textemdash\textemdash\textemdash\textemdash\textemdash\textemdash\qquad
         10
         \qquad\textemdash\textemdash\textemdash\textemdash\textemdash\textemdash\textemdash\textemdash} \\
    \bottomrule
    \end{tabular}}
\end{table}

\subsubsection{Coreset Selection} \label{sec:supp:coreset}
To compare the image-text retrieval performance of our distilled data with traditional coreset selection methods~\cite{agarwal2020contextual, coleman2019selection, ducoffe2018adversarial, iyer2021submodular, killamsetty2021grad, killamsetty2021glister, margatina2021active, mirzasoleiman2020coresets, paul2021deep, sener2017active, c-forget, c-herd, c-kcenter}, we selected several benchmarked methods as practiced in \cite{wu2024vldistill,xu2024lors}. Specifically, we reproduced the results using DeepCore~\cite{guo2022deepcore} for Herding~\cite{c-herd}, K-center~\cite{c-kcenter}, and Forgetting~\cite{c-forget} in Table \textcolor{cvprblue}{1} of our main paper.

\noindent\textbf{Herding}~\cite{c-herd}.\quad
We adapt DeepCore~\cite{guo2022deepcore}’s herding strategy to the image–text retrieval setting using a CLIP-style encoder. We first warm up the image encoder and text projection for $5$ epochs on the full training set with InfoNCE loss, using SGD (learning rate $0.1$, batch size $64$). After warmup epochs, we extract a joint representation for each pair by concatenating the $\ell_2$-normalized image and text features, and compute the global mean feature over all training pairs. Herding then greedily selects pairs whose features make the cumulative sum of selected features closest to $(t+1)$ times the global mean at step $t$, yielding an ordered coreset of all training pairs. For each budget ${\tilde B} \in \{100,200,500\}$, we take the first ${\tilde B}$ pairs in this ordering.

\noindent\textbf{K-center}~\cite{c-kcenter}.\quad
For K-center, we again use the same 5-epoch CLIP warmup (SGD optimizer, learning rate $0.1$, batch size $64$) and extract pairwise joint sum features. We then run greedy K-center on these embeddings: the first center is chosen uniformly at random, and subsequent centers are added by iteratively selecting the pair with the maximum Euclidean distance to the current selected set (farthest-first traversal) until we reach the maximum budget. 

\noindent\textbf{Forgetting}~\cite{c-forget}.\quad
For forgetting-based selection, we train the initial model on the full training set for $10$ epochs with InfoNCE loss, using an SGD optimizer. At each training epoch, we compute the CLIP similarity logits for every batch and evaluate whether each image–text pair is correctly retrieved. A pair is marked as ``correct" only when its image retrieves its own caption at rank-1 and, simultaneously, the caption retrieves its corresponding image at rank-1. For every pair, we maintain its correctness state over epochs and increment a forgetting counter whenever the pair transitions from correct to incorrect. We also track whether the pair was ever learned (\ie, correct at least once) and whether it remains correct in the final epoch. After $10$ epochs of training, we assign each pair a ranking score composed of its number of forgetting events and an additional penalty for pairs that were never learned. Sorting all pairs by this score in ascending order produces a forgetting-based coreset ordering, from which the first ${\tilde B}$ pairs define the selected subset.

\subsection{Performance Sensitivity}\label{sec:supp:sensitivity}

\subsubsection{Expert Pool Size}\label{sec:supp:expert_pool}

\begin{figure}[!t]
    \centering
    \includegraphics[width=.99\linewidth,trim={0.3cm 0 1cm 0},clip]{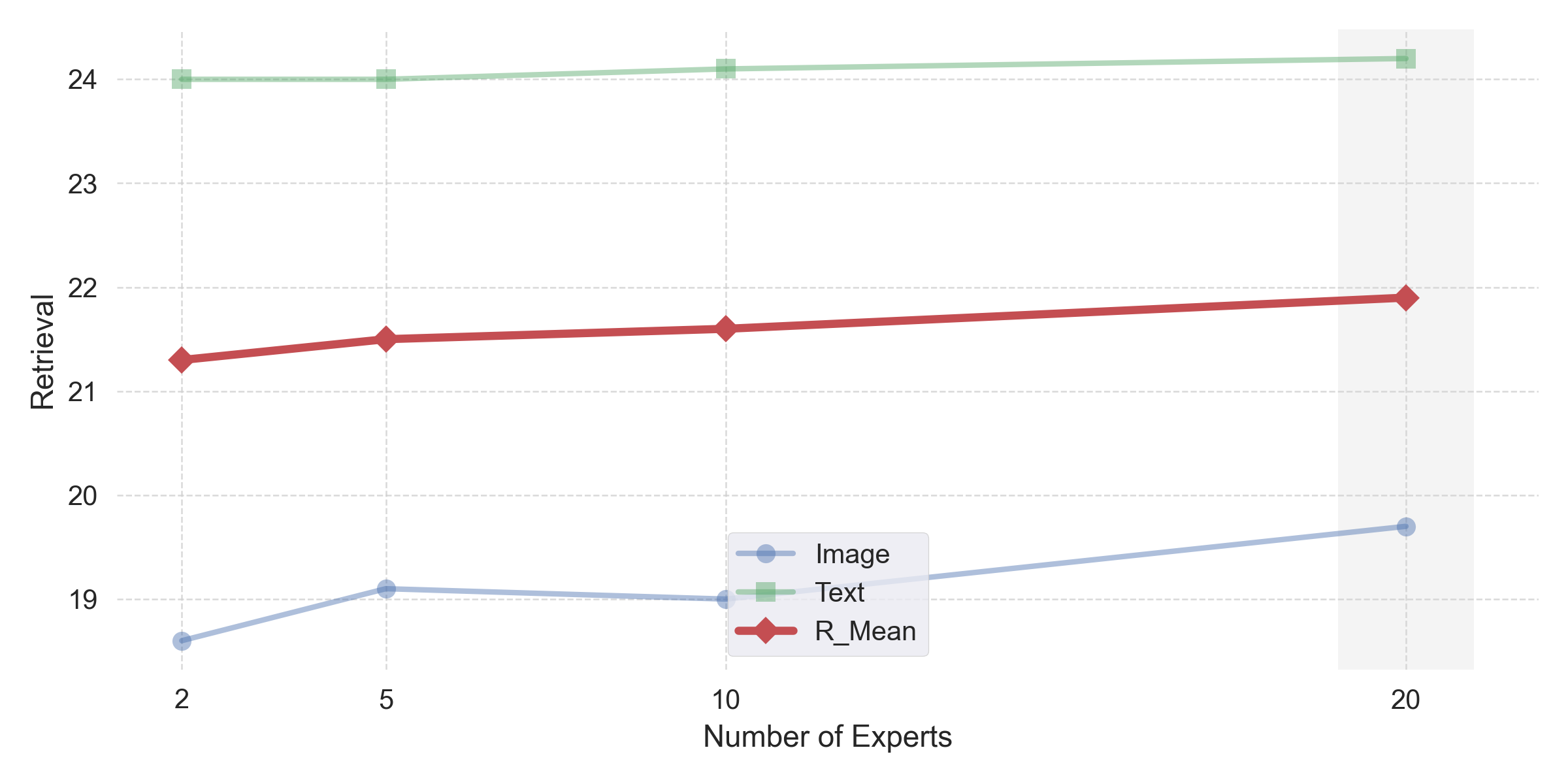}
    \vspace{-5mm}
    \caption{Retrieval performance on Flickr8k with 100 pairs as a function of the number of experts in the randomly sampled pool. Performance remains largely stable, with a slight improvement as the pool size increases.}
    \label{fig:supp:expert_pool}
    \vspace{-8pt}
\end{figure}

We further investigate how the size of the finetuned expert pool used for model initialization affects image–text retrieval. Specifically, we vary the number of available finetuned experts from which models are \textit{randomly sampled} (non-overlapping $N$ experts), and report retrieval performance across different pool sizes in Fig.~\ref{fig:supp:expert_pool}. The results indicate that performance remains consistently strong, with a slight improvement as the expert pool grows. We attribute this behavior to increased stability in random expert sampling when more experts are available, which in turn raises the likelihood of selecting a more diverse set of experts.

\subsubsection{Weighting Factors} \label{sec:supp:hyperparameter}
\begin{figure}[!ht]
    \centering
    \includegraphics[width=.99\linewidth]{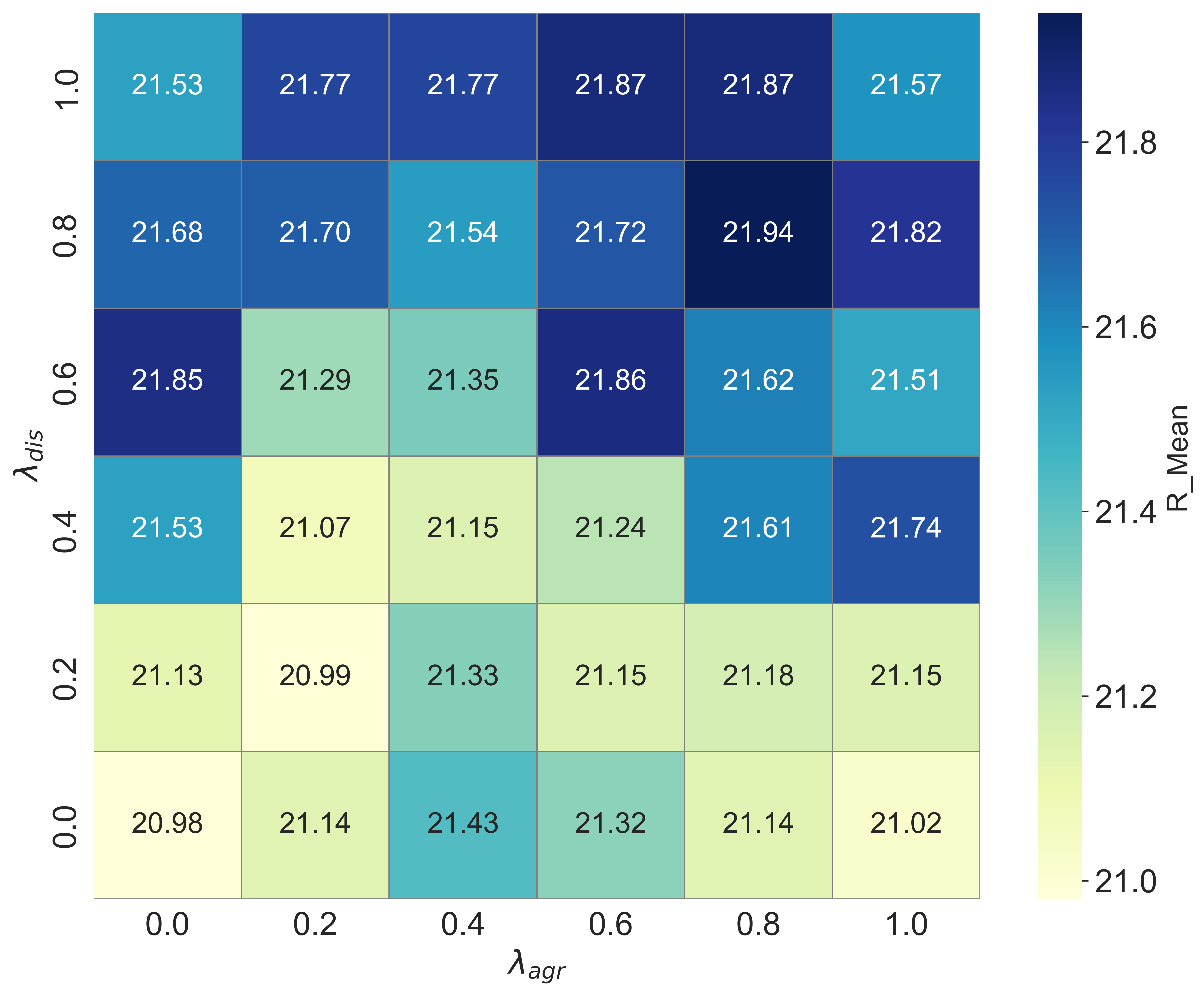}
    \vspace{-4mm}
    \caption{Hyperparameter sensitivity. While all choices of $\lambda_{agr}$ and $\lambda_{dis}$ consistently perform high, our selected values return the highest mean recall in image-text retrieval tasks.}
    \label{fig:supp:hyperparameter_sensitivity}
    \vspace{-8pt}
\end{figure}

\noindent\textbf{On hyperparameters $\lambda_{\mathrm{agr}}$} and \textbf{$\lambda_{\mathrm{dis}}$.}\quad
We tested the sensitivity of the hyperparameter to the weighting factors for the loss of agreement and discrepancy, \ie, $\lambda_{agr}$, and $\lambda_{dis}$ by sweeping the range $[0,1]$. As shown in Fig.~\ref{fig:supp:hyperparameter_sensitivity}, we choose the weighting factors $\lambda_{\mathrm{agr}}$ and $\lambda_{\mathrm{dis}}$ that yield the highest retrieval scores, at (0.8, 0.8). We observe that $\lambda_{\mathrm{dis}}$ contributes more to performance than $\lambda_{\mathrm{agr}}$ as there is more variation in performance with higher $\lambda_{\mathrm{dis}}$ relative to $\lambda_{\mathrm{agr}}$. Performance improvement is relatively marginal with $\lambda_{\mathrm{agr}}$ in all sweep values, which is consistent with the results of a slightly incremental improvement shown in Table \textcolor{cvprblue}{5} (ablation study).

\begin{figure}[!h]
    \centering
    \includegraphics[width=.99\linewidth]{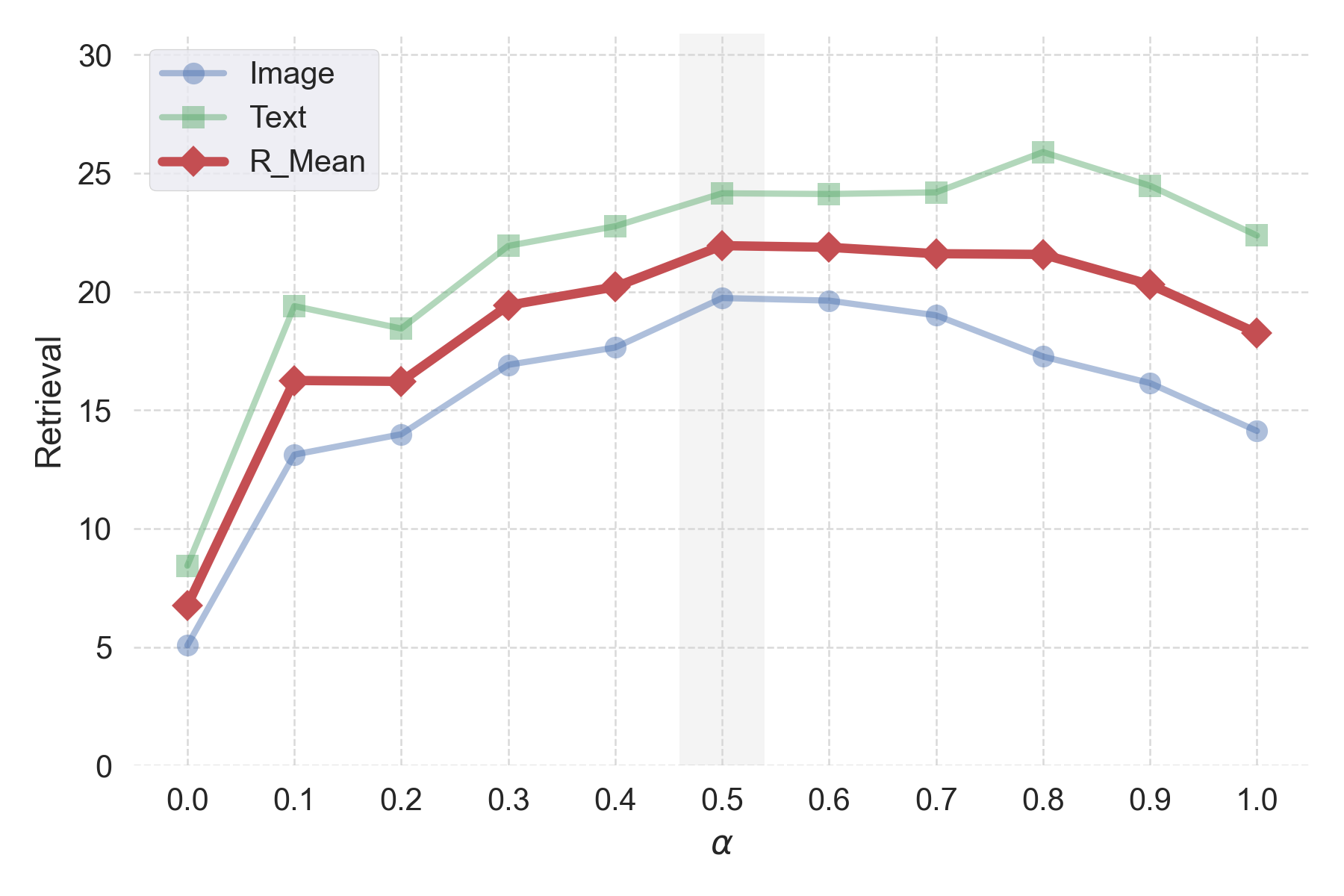}
    \vspace{-5mm}
    \caption{Hyperparameter sensitivity. While all choices of $\lambda_{agr}$ and $\lambda_{dis}$ consistently perform high, our selected values return the highest mean recall in image-text retrieval tasks.}
    \label{fig:supp:alpha}
    \vspace{-8pt}
\end{figure}
\begin{table*}[!h]
\caption{\textbf{Full Image-text retrieval results} for 100, 200, and 500 synthetic pairs using the coreset methods and distillation method, including standard deviation over five random evaluations. The condensation rate for \{Flickr8k, Flickr30k, and COCO\} datasets are approximately \{1.7\%, 0.3\%, 0.8\textperthousand\}, \{3.3\%, 0.7\%, 1.7\textperthousand\}, \{8.3\%, 1.7\%, 4.4\textperthousand\} for 100, 200, and 500 pairs. Best and runner-up results are indicated in \textbf{boldface} and \underline{underline}, respectively.}
\vspace{-2mm}
\label{tab:main_full}
\centering
\setlength{\tabcolsep}{0.5pt}
\renewcommand{\arraystretch}{1}
\renewcommand{\aboverulesep}{2pt}
\renewcommand{\belowrulesep}{2pt}
\vspace{-1mm}
\resizebox{\linewidth}{!}{%
\begin{tabular}{cl|ccccccc ccccccc ccccccc}
\toprule
\multirow{2}{*}{\rotatebox[origin=c]{90}{\# Pairs}}
& \multirow{1}{*}{\qquad\quad Dataset}
& \multicolumn{7}{c}{Flickr8k~\cite{hodosh2013flickr8k}} 
& \multicolumn{7}{c}{Flickr30k~\cite{young2014flickr}} 
& \multicolumn{7}{c}{COCO~\cite{COCO}} \\
\cmidrule(lr){3-9} \cmidrule(lr){10-16} \cmidrule(lr){17-23} 
& \multirow{1}{*}{Method}
& \textrm{IR@1} & \textrm{IR@5} & \textrm{IR@10} & \textrm{TR@1} & \textrm{TR@5} & \textrm{TR@10} & \cellcolor{Goldenrod!20} \textrm{Mean}
& \textrm{IR@1} & \textrm{IR@5} & \textrm{IR@10} & \textrm{TR@1} & \textrm{TR@5} & \textrm{TR@10} & \cellcolor{Goldenrod!20}\textrm{Mean}
& \textrm{IR@1} & \textrm{IR@5} & \textrm{IR@10} & \textrm{TR@1} & \textrm{TR@5} & \textrm{TR@10} & \cellcolor{Goldenrod!20}\textrm{Mean} \\
\midrule
\multirow{12}{*}{\rotatebox{90}{100}}
& Random 
& 1.2 & 5.6 & 9.6 & 2.7 & 8.0 & 12.6 & \cellcolor{Goldenrod!20}6.6
& 0.9 & 4.2 & 7.3 & 2.0 & 7.9 & 12.1 & \cellcolor{Goldenrod!20}5.7
& 0.4 & 1.7 & 3.0 & 0.8 & 3.2 & 5.4 & \cellcolor{Goldenrod!20}2.4 \\

& Herding~\cite{c-herd} 
& 1.2 & 4.4 & 8.5 & 2.2 & 8.5 & 14.2 & \cellcolor{Goldenrod!20}6.5
& 0.9 & 3.5 & 6.5 & 2.0 & 6.9 & 11.1 & \cellcolor{Goldenrod!20}5.1
& 0.3 & 1.4 & 2.6 & 0.8 & 3.0 & 5.5 & \cellcolor{Goldenrod!20}2.3 \\

& K-Center~\cite{c-kcenter} 
& 1.2 & 4.9 & 9.0 & 2.7 & 9.3 & 13.9 & \cellcolor{Goldenrod!20}6.8
& 1.1 & 4.9 & 8.7 & 3.0 & 9.1 & 14.3 & \cellcolor{Goldenrod!20}6.8
& 0.5 & 1.9 & 3.6 & 1.1 & 4.2 & 7.6 & \cellcolor{Goldenrod!20}3.2 \\

& Forgetting~\cite{c-forget} 
& 1.2 & 4.1 & 7.1 & 1.5 & 4.8 & 8.4 & \cellcolor{Goldenrod!20}4.5
& 0.8 & 3.6 & 6.2 & 1.2 & 5.4 & 9.1 & \cellcolor{Goldenrod!20}4.4
& 0.2 & 1.0 & 1.9 & 0.2 & 1.2 & 2.6 & \cellcolor{Goldenrod!20}1.2 \\

\cdashline{2-23}

& MTT-VL~\cite{wu2024vldistill} 
& 0.8 & 4.1 & 7.0 & 1.2 & 6.4 & 11.5 & \cellcolor{Goldenrod!20}5.1
& 4.7 & 15.7 & 24.6 & 9.9 & 28.3 & 39.1 & \cellcolor{Goldenrod!20}20.4
& 1.3 & 5.4 & 9.5 & 2.5 & 10.0 & 15.7 & \cellcolor{Goldenrod!20}7.4 \\
&  & \footnotesize{\textcolor{gray}{$\pm$ 0.0}} & \footnotesize{\textcolor{gray}{$\pm$ 0.2}} & \footnotesize{\textcolor{gray}{$\pm$ 0.3}} & \footnotesize{\textcolor{gray}{$\pm$ 0.2}} & \footnotesize{\textcolor{gray}{$\pm$ 0.6}} & \footnotesize{\textcolor{gray}{$\pm$ 0.9}} 
& \cellcolor{Goldenrod!20}\footnotesize{\textcolor{gray}{$\pm$ 0.3}} & \footnotesize{\textcolor{gray}{$\pm$ 0.2}} & \footnotesize{\textcolor{gray}{$\pm$ 0.5}} & \footnotesize{\textcolor{gray}{$\pm$ 1.0}} & \footnotesize{\textcolor{gray}{$\pm$ 0.3}} & \footnotesize{\textcolor{gray}{$\pm$ 0.5}} & \footnotesize{\textcolor{gray}{$\pm$ 0.7}} 
& \cellcolor{Goldenrod!20}\footnotesize{\textcolor{gray}{$\pm$ 0.5}} & \footnotesize{\textcolor{gray}{$\pm$ 0.1}} & \footnotesize{\textcolor{gray}{$\pm$ 0.3}} & \footnotesize{\textcolor{gray}{$\pm$ 0.5}} & \footnotesize{\textcolor{gray}{$\pm$ 0.3}} & \footnotesize{\textcolor{gray}{$\pm$ 0.5}} & \footnotesize{\textcolor{gray}{$\pm$ 0.4}}
& \cellcolor{Goldenrod!20}\footnotesize{\textcolor{gray}{$\pm$ 0.4}} \\

& TESLA$_{\textrm{WBCE}}$
& 0.8 & 3.8 & 7.0 & 4.7 & 16.1 & 25.9 & \cellcolor{Goldenrod!20} 9.7 
& 0.5 & 2.3 & 4.7 & 5.5 & 19.5 & 28.9 & \cellcolor{Goldenrod!20}10.2
& 0.3 & 1.0 & 1.8 & 2.0 & 7.7 & 13.5 & \cellcolor{Goldenrod!20}4.4 \\
&  & \footnotesize{\textcolor{gray}{$\pm$ 0.1}} & \footnotesize{\textcolor{gray}{$\pm$ 0.4}} & \footnotesize{\textcolor{gray}{$\pm$ 0.3}} & \footnotesize{\textcolor{gray}{$\pm$ 0.3}} & \footnotesize{\textcolor{gray}{$\pm$ 0.6}} & \footnotesize{\textcolor{gray}{$\pm$ 0.8}} 
& \cellcolor{Goldenrod!20}\footnotesize{\textcolor{gray}{$\pm$ 0.3}} 
& \footnotesize{\textcolor{gray}{$\pm$ 0.2}} & \footnotesize{\textcolor{gray}{$\pm$ 0.2}} & \footnotesize{\textcolor{gray}{$\pm$ 0.4}} & \footnotesize{\textcolor{gray}{$\pm$ 0.5}} & \footnotesize{\textcolor{gray}{$\pm$ 0.9}} & \footnotesize{\textcolor{gray}{$\pm$ 1.0}} 
& \cellcolor{Goldenrod!20}\footnotesize{\textcolor{gray}{$\pm$ 0.5}} 
& \footnotesize{\textcolor{gray}{$\pm$ 0.2}} & \footnotesize{\textcolor{gray}{$\pm$ 0.4}} & \footnotesize{\textcolor{gray}{$\pm$ 0.5}} & \footnotesize{\textcolor{gray}{$\pm$ 0.2}} & \footnotesize{\textcolor{gray}{$\pm$ 0.3}} & \footnotesize{\textcolor{gray}{$\pm$ 0.3}} & \cellcolor{Goldenrod!20}\footnotesize{\textcolor{gray}{$\pm$ 0.4}} \\

& LoRS~\cite{xu2024lors} 
& 4.9 & 18.0 & 29.0 & 7.0 & 22.8 & 34.8 & \cellcolor{Goldenrod!20}\underline{19.4}
& 8.3 & 24.1 & 35.1 & 11.8 & 35.8 & 49.2 & \cellcolor{Goldenrod!20}\textbf{27.4}
& 1.8 & 7.1 & 12.2 & 3.3 & 12.2 & 19.6 & \cellcolor{Goldenrod!20}\underline{9.4} \\
&  & \footnotesize{\textcolor{gray}{$\pm$ 0.3}} & \footnotesize{\textcolor{gray}{$\pm$ 0.8}} & \footnotesize{\textcolor{gray}{$\pm$ 1.2}} & \footnotesize{\textcolor{gray}{$\pm$ 0.3}} & \footnotesize{\textcolor{gray}{$\pm$ 0.4}} & \footnotesize{\textcolor{gray}{$\pm$ 0.8}} & \cellcolor{Goldenrod!20}\footnotesize{\textcolor{gray}{$\pm$ 0.5}} 
& \footnotesize{\textcolor{gray}{$\pm$ 0.2}} & \footnotesize{\textcolor{gray}{$\pm$ 0.2}} & \footnotesize{\textcolor{gray}{$\pm$ 0.3}} & \footnotesize{\textcolor{gray}{$\pm$ 0.2}} & \footnotesize{\textcolor{gray}{$\pm$ 0.6}} & \footnotesize{\textcolor{gray}{$\pm$ 0.5}} & \cellcolor{Goldenrod!20}\footnotesize{\textcolor{gray}{$\pm$ 0.3}} 
& \footnotesize{\textcolor{gray}{$\pm$ 0.1}} & \footnotesize{\textcolor{gray}{$\pm$ 0.2}} & \footnotesize{\textcolor{gray}{$\pm$ 2.0}} & \footnotesize{\textcolor{gray}{$\pm$ 0.2}} & \footnotesize{\textcolor{gray}{$\pm$ 0.3}} & \footnotesize{\textcolor{gray}{$\pm$ 0.3}} & \cellcolor{Goldenrod!20}\footnotesize{\textcolor{gray}{$\pm$ 0.5}} \\

& \cellcolor{Goldenrod!20}Ours
& 6.0 & 20.8 & 32.4 & 7.9 & 26.5 & 38.1          &\cellcolor{Goldenrod!20}\textbf{21.9} 
&8.1	&24.7	&36.2&	11.5&	32.6&	45.0	& \cellcolor{Goldenrod!20}\underline{26.4}
& 1.9 &  7.6 & 13.2   &    3.6 & 13.7 &  21.6  & \cellcolor{Goldenrod!20}\textbf{10.3} \\
&  & \footnotesize{\textcolor{gray}{$\pm$ 0.4}} & \footnotesize{\textcolor{gray}{$\pm$ 0.4}} & \footnotesize{\textcolor{gray}{$\pm$ 0.7}} & \footnotesize{\textcolor{gray}{$\pm$ 0.3}} & \footnotesize{\textcolor{gray}{$\pm$ 0.5}} & \footnotesize{\textcolor{gray}{$\pm$ 0.6}} & \cellcolor{Goldenrod!20}\footnotesize{\textcolor{gray}{$\pm$ 0.3}} 
& \footnotesize{\textcolor{gray}{$\pm$ 0.3}} & \footnotesize{\textcolor{gray}{$\pm$ 0.8}} & \footnotesize{\textcolor{gray}{$\pm$ 0.7}} & \footnotesize{\textcolor{gray}{$\pm$ 0.7}} & \footnotesize{\textcolor{gray}{$\pm$ 0.9}} & \footnotesize{\textcolor{gray}{$\pm$ 1.0}} & \cellcolor{Goldenrod!20}\footnotesize{\textcolor{gray}{$\pm$ 0.3}} 
& \footnotesize{\textcolor{gray}{$\pm$ 0.1}} & \footnotesize{\textcolor{gray}{$\pm$ 0.1}} & \footnotesize{\textcolor{gray}{$\pm$ 0.2}} & \footnotesize{\textcolor{gray}{$\pm$ 0.2}} & \footnotesize{\textcolor{gray}{$\pm$ 0.3}} & \footnotesize{\textcolor{gray}{$\pm$ 0.4}} & \cellcolor{Goldenrod!20}\footnotesize{\textcolor{gray}{$\pm$ 0.2}} \\

\midrule

\multirow{12}{*}{\rotatebox{90}{200}}
& Random 
& 2.0 & 7.8 & 13.7 & 3.3 & 12.5 & 19.5 & \cellcolor{Goldenrod!20}9.8
& 1.9 & 7.1 & 12.3 & 1.9 & 10.3 & 18.2 & \cellcolor{Goldenrod!20}8.6
& 0.6 & 2.7 & 4.9 & 1.3 & 5.3 & 9.0 & \cellcolor{Goldenrod!20}4.0 \\

& Herding~\cite{c-herd} 
& 2.0 & 7.6 & 14.0 & 3.2 & 12.5 & 19.9 & \cellcolor{Goldenrod!20}9.9
& 1.4 & 5.9 & 10.5 & 3.1 & 9.4 & 15.5 &\cellcolor{Goldenrod!20} 7.6
& 0.6 & 2.5 & 4.6 & 1.1 & 4.6 & 8.4 & \cellcolor{Goldenrod!20}3.6\\

& K-Center~\cite{c-kcenter} 
& 2.3 & 9.1 & 15.0 & 3.8 & 13.7 & 20.9 & \cellcolor{Goldenrod!20} 10.8
& 2.2 & 8.1 & 13.3 & 4.2 & 13.1 & 21.2 & \cellcolor{Goldenrod!20}10.3
& 0.9 & 3.4 & 5.9 & 2.1 & 7.0 & 11.6 & \cellcolor{Goldenrod!20}5.1 \\

& Forgetting~\cite{c-forget} 
& 1.7 & 6.5 & 11.5 & 3.1 & 9.7 & 15.4 & \cellcolor{Goldenrod!20}8.0
& 1.6 & 6.6 & 10.8 & 2.5 & 9.0 & 14.9 & \cellcolor{Goldenrod!20}7.6
& 0.4 & 1.6 & 3.0 & 0.7 & 2.8 & 5.1 & \cellcolor{Goldenrod!20}2.3 \\

\cdashline{2-23}

& MTT-VL~\cite{wu2024vldistill} 
& 1.8 & 7.0 & 12.2 & 2.8 & 10.3 & 17.3 & \cellcolor{Goldenrod!20}8.6
& 4.6 & 16.0 & 25.5 & 10.2 & 28.7 & 41.9 & \cellcolor{Goldenrod!20}21.2
& 1.7 & 6.5 & 12.3 & 3.3 & 11.9 & 19.4 & \cellcolor{Goldenrod!20}9.2 \\
& & \footnotesize{\textcolor{gray}{$\pm$ 0.2}} & \footnotesize{\textcolor{gray}{$\pm$ 0.2}} & \footnotesize{\textcolor{gray}{$\pm$ 0.2}} & \footnotesize{\textcolor{gray}{$\pm$ 0.3}} & \footnotesize{\textcolor{gray}{$\pm$ 0.7}} & \footnotesize{\textcolor{gray}{$\pm$ 0.7}} & \cellcolor{Goldenrod!20}\footnotesize{\textcolor{gray}{$\pm$ 0.3}} 
& \footnotesize{\textcolor{gray}{$\pm$ 0.9}} & \footnotesize{\textcolor{gray}{$\pm$ 1.6}} & \footnotesize{\textcolor{gray}{$\pm$ 2.6}} & \footnotesize{\textcolor{gray}{$\pm$ 0.8}} & \footnotesize{\textcolor{gray}{$\pm$ 1.0}} & \footnotesize{\textcolor{gray}{$\pm$ 1.9}} & \cellcolor{Goldenrod!20}\footnotesize{\textcolor{gray}{$\pm$ 1.5}} 
& \footnotesize{\textcolor{gray}{$\pm$ 0.1}} & \footnotesize{\textcolor{gray}{$\pm$ 0.4}} & \footnotesize{\textcolor{gray}{$\pm$ 0.8}} & \footnotesize{\textcolor{gray}{$\pm$ 0.2}} & \footnotesize{\textcolor{gray}{$\pm$ 0.6}} & \footnotesize{\textcolor{gray}{$\pm$ 1.2}} & \cellcolor{Goldenrod!20}\footnotesize{\textcolor{gray}{$\pm$ 0.6}} \\

& TESLA$_{\textrm{WBCE}}$
& 1.2 & 4.7 & 8.4 & 6.6 & 19.5 & 29.5 & \cellcolor{Goldenrod!20}11.7
& 0.2 & 1.3 & 2.5 & 2.8 & 10.4 & 17.4 & \cellcolor{Goldenrod!20}5.8
& 0.1 & 0.2 & 0.5 & 0.7 & 3.1 & 5.3 & \cellcolor{Goldenrod!20}1.7 \\
&  & \footnotesize{\textcolor{gray}{$\pm$ 0.2}} & \footnotesize{\textcolor{gray}{$\pm$ 0.5}} & \footnotesize{\textcolor{gray}{$\pm$ 0.6}} & \footnotesize{\textcolor{gray}{$\pm$ 0.3}} & \footnotesize{\textcolor{gray}{$\pm$ 1.1}} & \footnotesize{\textcolor{gray}{$\pm$ 1.5}} & \cellcolor{Goldenrod!20}\footnotesize{\textcolor{gray}{$\pm$ 0.6}} 
& \footnotesize{\textcolor{gray}{$\pm$ 0.1}} & \footnotesize{\textcolor{gray}{$\pm$ 0.2}} & \footnotesize{\textcolor{gray}{$\pm$ 0.2}} & \footnotesize{\textcolor{gray}{$\pm$ 0.5}} & \footnotesize{\textcolor{gray}{$\pm$ 1.5}} & \footnotesize{\textcolor{gray}{$\pm$ 1.6}} & \cellcolor{Goldenrod!20}\footnotesize{\textcolor{gray}{$\pm$ 0.7}} 
& \footnotesize{\textcolor{gray}{$\pm$ 0.1}} & \footnotesize{\textcolor{gray}{$\pm$ 0.1}} & \footnotesize{\textcolor{gray}{$\pm$ 0.1}} & \footnotesize{\textcolor{gray}{$\pm$ 0.2}} & \footnotesize{\textcolor{gray}{$\pm$ 0.5}} & \footnotesize{\textcolor{gray}{$\pm$ 0.8}} & \cellcolor{Goldenrod!20}\footnotesize{\textcolor{gray}{$\pm$ 0.3}} \\

& LoRS~\cite{xu2024lors} 
& 6.3 & 20.5 & 31.6 & 9.5 & 26.3 & 38.2 & \cellcolor{Goldenrod!20}\underline{22.1}
& 8.6 & 25.3 & 36.6 & 14.5 & 38.7 & 53.4 & \cellcolor{Goldenrod!20}\textbf{29.5}
& 2.4 & 9.3 & 15.5 & 4.3 & 14.2 & 22.6 & \cellcolor{Goldenrod!20}\underline{11.4} \\
&  & \footnotesize{\textcolor{gray}{$\pm$ 0.4}} & \footnotesize{\textcolor{gray}{$\pm$ 0.7}} & \footnotesize{\textcolor{gray}{$\pm$ 0.7}} & \footnotesize{\textcolor{gray}{$\pm$ 0.4}} & \footnotesize{\textcolor{gray}{$\pm$ 0.4}} & \footnotesize{\textcolor{gray}{$\pm$ 0.9}} & \cellcolor{Goldenrod!20}\footnotesize{\textcolor{gray}{$\pm$ 0.4}} 
& \footnotesize{\textcolor{gray}{$\pm$ 0.3}} & \footnotesize{\textcolor{gray}{$\pm$ 0.2}} & \footnotesize{\textcolor{gray}{$\pm$ 0.3}} & \footnotesize{\textcolor{gray}{$\pm$ 0.5}} & \footnotesize{\textcolor{gray}{$\pm$ 0.5}} &  \footnotesize{\textcolor{gray}{$\pm$ 0.5}} & \cellcolor{Goldenrod!20}\footnotesize{\textcolor{gray}{$\pm$ 0.4}} 
& \footnotesize{\textcolor{gray}{$\pm$ 0.1}} & \footnotesize{\textcolor{gray}{$\pm$ 0.2}} & \footnotesize{\textcolor{gray}{$\pm$ 0.2}} & \footnotesize{\textcolor{gray}{$\pm$ 0.1}} & \footnotesize{\textcolor{gray}{$\pm$ 0.3}} & \footnotesize{\textcolor{gray}{$\pm$ 0.2}} & \cellcolor{Goldenrod!20}\footnotesize{\textcolor{gray}{$\pm$ 0.2}}
 \\

& \cellcolor{Goldenrod!20}Ours
& 7.1	&23.2&	35.1	&	9.9	&29.0	&41.6	& \cellcolor{Goldenrod!20}\textbf{24.3}
& 9.1&	26.7&	39.1	&	13.0&	33.7&	47.4&	\cellcolor{Goldenrod!20}\underline{28.2} 
& 2.9  &  11.1  &  18.4 &   4.9  &  16.2  &  25.3   &\cellcolor{Goldenrod!20}\textbf{13.1} \\
&  & \footnotesize{\textcolor{gray}{$\pm$ 0.2}} & \footnotesize{\textcolor{gray}{$\pm$ 0.4}} & \footnotesize{\textcolor{gray}{$\pm$ 0.6}} & \footnotesize{\textcolor{gray}{$\pm$ 0.2}} & \footnotesize{\textcolor{gray}{$\pm$ 0.7}} & \footnotesize{\textcolor{gray}{$\pm$ 0.6}} & \cellcolor{Goldenrod!20}\footnotesize{\textcolor{gray}{$\pm$ 0.3}} 
& \footnotesize{\textcolor{gray}{$\pm$ 0.2}} & \footnotesize{\textcolor{gray}{$\pm$ 0.4}} & \footnotesize{\textcolor{gray}{$\pm$ 0.4}} & \footnotesize{\textcolor{gray}{$\pm$ 0.5}} & \footnotesize{\textcolor{gray}{$\pm$ 0.9}} & \footnotesize{\textcolor{gray}{$\pm$ 0.5}} & \cellcolor{Goldenrod!20}\footnotesize{\textcolor{gray}{$\pm$ 0.2}} 
& \footnotesize{\textcolor{gray}{$\pm$ 0.1}} & \footnotesize{\textcolor{gray}{$\pm$ 0.2}} & \footnotesize{\textcolor{gray}{$\pm$ 0.3}} & \footnotesize{\textcolor{gray}{$\pm$ 0.1}} & \footnotesize{\textcolor{gray}{$\pm$ 0.3}} & \footnotesize{\textcolor{gray}{$\pm$ 0.3}} & \cellcolor{Goldenrod!20}\footnotesize{\textcolor{gray}{$\pm$ 0.1}} \\

\midrule

\multirow{12}{*}{\rotatebox{90}{500}}
& Random 
& 3.7 & 13.0 & 21.2 & 6.0 & 19.4 & 28.8 & \cellcolor{Goldenrod!20}15.3
& 3.2 & 11.5 & 18.9 & 5.2 & 18.3 & 27.4 &\cellcolor{Goldenrod!20} 14.1
& 1.2 & 5.2 & 9.2 & 2.5 & 8.7 & 14.9 & \cellcolor{Goldenrod!20}7.0\\

& Herding~\cite{c-herd} 
& 3.7 & 12.5 & 19.8 & 4.9 & 17.5 & 26.4 & \cellcolor{Goldenrod!20}14.1
& 2.7 & 10.6 & 17.0 & 4.1 & 14.9 & 24.0 & \cellcolor{Goldenrod!20}12.2
& 1.3 & 5.0 & 8.8 & 2.0 & 7.9 & 13.6 & \cellcolor{Goldenrod!20}6.4\\

& K-Center~\cite{c-kcenter} 
& 4.0 & 13.4 & 21.1 & 5.9 & 18.9 & 29.0 & \cellcolor{Goldenrod!20}15.4
& 3.4 & 11.8 & 18.7 & 6.7 & 18.0 & 30.6 & \cellcolor{Goldenrod!20}14.9
& 1.5 & 5.7 & 9.7 & 3.0 & 9.9 & 16.2 & \cellcolor{Goldenrod!20}7.7\\

& Forgetting~\cite{c-forget} 
& 4.6 & 16.2 & 24.5 & 5.8 & 21.7 & 31.7 & \cellcolor{Goldenrod!20}17.4
& 3.6 & 12.7 & 20.6 & 6.1 & 18.7 & 29.5 & \cellcolor{Goldenrod!20}15.2
& 1.1 & 4.3 & 7.6 & 2.0 & 7.3 & 11.3 & \cellcolor{Goldenrod!20}5.6 \\

\cdashline{2-23}

& MTT-VL~\cite{wu2024vldistill} 
& 3.7 & 13.3 & 21.3 & 5.8 & 18.0 & 28.2 & \cellcolor{Goldenrod!20}15.1
& 6.6 & 20.2 & 30.0 & 13.3 & 32.8 & 46.8 & \cellcolor{Goldenrod!20}25.0
& 2.5 & 8.9 & 15.8 & 5.0 & 17.2 & 26.0 & \cellcolor{Goldenrod!20}12.6 \\
&  & \footnotesize{\textcolor{gray}{$\pm$ 0.0}} & \footnotesize{\textcolor{gray}{$\pm$ 0.3}} & \footnotesize{\textcolor{gray}{$\pm$ 0.5}} & \footnotesize{\textcolor{gray}{$\pm$ 0.3}} & \footnotesize{\textcolor{gray}{$\pm$ 0.6}} & \footnotesize{\textcolor{gray}{$\pm$ 0.7}} & \cellcolor{Goldenrod!20}\footnotesize{\textcolor{gray}{$\pm$ 0.3}} 
& \footnotesize{\textcolor{gray}{$\pm$ 0.3}} & \footnotesize{\textcolor{gray}{$\pm$ 1.2}} & \footnotesize{\textcolor{gray}{$\pm$ 2.1}} & \footnotesize{\textcolor{gray}{$\pm$ 0.6}} & \footnotesize{\textcolor{gray}{$\pm$ 1.8}} & \footnotesize{\textcolor{gray}{$\pm$ 0.8}} & \cellcolor{Goldenrod!20}\footnotesize{\textcolor{gray}{$\pm$ 1.1}} 
& \footnotesize{\textcolor{gray}{$\pm$ 0.5}} & \footnotesize{\textcolor{gray}{$\pm$ 0.7}} & \footnotesize{\textcolor{gray}{$\pm$ 1.5}} & \footnotesize{\textcolor{gray}{$\pm$ 0.4}} & \footnotesize{\textcolor{gray}{$\pm$ 1.3}} & \footnotesize{\textcolor{gray}{$\pm$ 1.9}} & \cellcolor{Goldenrod!20}\footnotesize{\textcolor{gray}{$\pm$ 1.1}} \\

&
TESLA$_{\textrm{WBCE}}$
& 2.5 & 8.8 & 14.1 & 6.9 & 19.6 & 29.0 & \cellcolor{Goldenrod!20}13.5
& 1.1 & 7.3 & 12.6 & 5.1 & 15.3 & 23.8 & \cellcolor{Goldenrod!20}10.9
& 0.8 & 3.6 & 6.7 & 1.7 & 5.9 & 10.2 & \cellcolor{Goldenrod!20}4.8\\
&  & \footnotesize{\textcolor{gray}{$\pm$ 0.2}} & \footnotesize{\textcolor{gray}{$\pm$ 0.3}} & \footnotesize{\textcolor{gray}{$\pm$ 0.2}} & \footnotesize{\textcolor{gray}{$\pm$ 0.4}} & \footnotesize{\textcolor{gray}{$\pm$ 0.6}} & \footnotesize{\textcolor{gray}{$\pm$ 0.6}} & \cellcolor{Goldenrod!20} \footnotesize{\textcolor{gray}{$\pm$ 0.2}} 
& \footnotesize{\textcolor{gray}{$\pm$ 0.2}} & \footnotesize{\textcolor{gray}{$\pm$ 0.4}} & \footnotesize{\textcolor{gray}{$\pm$ 0.5}} & \footnotesize{\textcolor{gray}{$\pm$ 0.2}} & \footnotesize{\textcolor{gray}{$\pm$ 0.5}} & \footnotesize{\textcolor{gray}{$\pm$ 0.3}} & \cellcolor{Goldenrod!20} \footnotesize{\textcolor{gray}{$\pm$ 0.4}} 
&  \footnotesize{\textcolor{gray}{$\pm$ 0.2}} &  \footnotesize{\textcolor{gray}{$\pm$ 0.6}} & \footnotesize{\textcolor{gray}{$\pm$ 0.9}} & \footnotesize{\textcolor{gray}{$\pm$ 0.4}} & \footnotesize{\textcolor{gray}{$\pm$ 0.8}} & \footnotesize{\textcolor{gray}{$\pm$ 1.0}} & \cellcolor{Goldenrod!20} \footnotesize{\textcolor{gray}{$\pm$ 0.7}} \\

&  LoRS~\cite{xu2024lors} 
& 6.9 & 22.0 & 33.1 & 10.9 & 31.0 & 45.8 & \cellcolor{Goldenrod!20}\underline{25.0}
& 10.0 & 28.9 & 41.6 & 15.5 & 29.8 & 53.7 & \cellcolor{Goldenrod!20}\textbf{31.6}
& 2.8 & 9.9 & 16.5 & 5.3 & 18.3 & 27.9 & \cellcolor{Goldenrod!20}\underline{13.5}\\
&  & \footnotesize{\textcolor{gray}{$\pm$ 0.4}} & \footnotesize{\textcolor{gray}{$\pm$ 1.2}} & \footnotesize{\textcolor{gray}{$\pm$ 1.6}} & \footnotesize{\textcolor{gray}{$\pm$ 0.3}} & \footnotesize{\textcolor{gray}{$\pm$ 1.0}} & \footnotesize{\textcolor{gray}{$\pm$ 1.2}} & \cellcolor{Goldenrod!20}\footnotesize{\textcolor{gray}{$\pm$ 0.7}} 
& \footnotesize{\textcolor{gray}{$\pm$ 0.2}} & \footnotesize{\textcolor{gray}{$\pm$ 0.7}} & \footnotesize{\textcolor{gray}{$\pm$ 0.6}} & \footnotesize{\textcolor{gray}{$\pm$ 0.7}} & \footnotesize{\textcolor{gray}{$\pm$ 0.4}} & \footnotesize{\textcolor{gray}{$\pm$ 0.3}} & \cellcolor{Goldenrod!20}\footnotesize{\textcolor{gray}{$\pm$ 0.5}} 
& \footnotesize{\textcolor{gray}{$\pm$ 0.2}} & \footnotesize{\textcolor{gray}{$\pm$ 0.5}} & \footnotesize{\textcolor{gray}{$\pm$ 0.7}} & \footnotesize{\textcolor{gray}{$\pm$ 0.5}} & \footnotesize{\textcolor{gray}{$\pm$ 1.5}} & \footnotesize{\textcolor{gray}{$\pm$ 1.4}} & \cellcolor{Goldenrod!20}\footnotesize{\textcolor{gray}{$\pm$ 0.8}} \\
&
\cellcolor{Goldenrod!20} Ours 
& 7.4 & 25.0 & 37.1 & 11.2 & 32.4 & 44.2 &\cellcolor{Goldenrod!20}\textbf{26.2}
& 10.0	& 29.3	& 42.0 &	13.7&	37.0	&51.5&	\cellcolor{Goldenrod!20}\underline{30.6}
&3.7	& 13.6 &	22.2	&5.6&	18.4	& 28.2 & \cellcolor{Goldenrod!20}\textbf{15.3} \\
&  & \footnotesize{\textcolor{gray}{$\pm$ 0.4}} & \footnotesize{\textcolor{gray}{$\pm$ 0.4}} & \footnotesize{\textcolor{gray}{$\pm$ 0.7}} & \footnotesize{\textcolor{gray}{$\pm$ 0.7}} & \footnotesize{\textcolor{gray}{$\pm$ 0.7}} & \footnotesize{\textcolor{gray}{$\pm$ 0.5}} & \cellcolor{Goldenrod!20}\footnotesize{\textcolor{gray}{$\pm$ 0.3}} 
& \footnotesize{\textcolor{gray}{$\pm$ 0.5}} & \footnotesize{\textcolor{gray}{$\pm$ 0.5}} & \footnotesize{\textcolor{gray}{$\pm$ 0.7}} & \footnotesize{\textcolor{gray}{$\pm$ 0.5}} & \footnotesize{\textcolor{gray}{$\pm$ 0.6}} & \footnotesize{\textcolor{gray}{$\pm$ 0.9}} & \cellcolor{Goldenrod!20}\footnotesize{\textcolor{gray}{$\pm$ 0.4}} 
& \footnotesize{\textcolor{gray}{$\pm$ 0.1}} & \footnotesize{\textcolor{gray}{$\pm$ 0.2}} & \footnotesize{\textcolor{gray}{$\pm$ 0.5}} & \footnotesize{\textcolor{gray}{$\pm$ 0.3}} & \footnotesize{\textcolor{gray}{$\pm$ 0.2}} & \footnotesize{\textcolor{gray}{$\pm$ 0.4}} & \cellcolor{Goldenrod!20}\footnotesize{\textcolor{gray}{$\pm$ 0.2}} \\

\midrule

& Full Dataset
& 25.5 & 56.1 & 69.2 & 32.7 & 64.5 & 74.5 & \cellcolor{Goldenrod!20}53.8 
& 28.1 & 57.9 & 70.3 & 34.9 & 65.4 & 77.6 & \cellcolor{Goldenrod!20}55.7
& 17.3 & 42.8 & 56.7 & 20.4 & 47.7 & 62.1 & \cellcolor{Goldenrod!20}41.2 \\

\bottomrule
\end{tabular}%
}
\end{table*}
\noindent\textbf{On hyperparameter $\alpha$.}\quad
The hyperparameter $\alpha$ in Eq. \textcolor{cvprblue}{5} of the main paper acts as a global scaling factor on the layer-wise mixing coefficient $t_{\ell}^{m}$, which determines how strongly each layer of the pretrained anchor $\theta_{0,\ell}^{m}$ is nudged toward the averaged finetuned updates $\tfrac{1}{2}(\Delta_{1,\ell}^{m}+\Delta_{2,\ell}^{m})$. When $\alpha=0$, the initialization collapses to the original pretrained model, whereas larger $\alpha$ values increasingly inject real-data finetuning information along the direction prescribed by $t_{\ell}^{m}$. As shown in Fig.~\ref{fig:supp:alpha}, retrieval performance rapidly improves as we move away from $\alpha=0$, peaks in an intermediate range, and then slightly degrades when $\alpha$ approaches $1.0$, where the model becomes overly biased toward the finetuned solutions. This behavior suggests that too small $\alpha$ under-exploits the benefits of real-data finetuning, while too large $\alpha$ over-specializes the distilled initialization. A mid-range value of $\alpha=0.5$, as adopted in ours, provides a balanced compromise and yields the best mean recall across image-to-text and text-to-image retrieval.

\subsubsection{Additional Insights}
\noindent\textbf{Weight mixing for large heterogeneous encoders.}\quad
We agree weight-space update-agreement should be scale-robust. \cite{jang2024modelstock} provides encouraging evidence in CLIP-scale models that fine-tuning updates exhibit highly structured, layer-wise regularities: displacement magnitudes and directions remain consistent across a wide range of fine-tuning conditions, and similar patterns are observed across backbone families (\eg, ViT, ResNet, ConvNeXt). Given these coherent patterns, using directional agreement as a conservative mixing criterion is well-motivated, since it explicitly checks the compatibility of updates rather than relying on unstructured averaging, which can introduce interference. 

\noindent\textbf{Comparison to a concurrent work.}
In comparison to a concurrent work, EDGE~\cite{zhao2025edge}, on the total cost, our pipeline includes a one-time expert training ($\sim$91.7h), whereas EDGE trains SDv1.5 ($\sim$49.0h) with additional caption generation time (\eg, $\sim$248.14s for 100 pairs); $^{\textcolor{red}{\ast}}$denotes measurement on an RTX A5000~\cite{zhao2025edge}, otherwise on an RTX A6000. 
Our end-to-end GPUh is slightly lower on Flickr30k but becomes higher on \texttt{COCO} as scale increases due to expert training, highlighting the limitation of expert-based methods.
\begin{table}[!h]
    \centering
    \vspace{-3.8mm}
    \centering
    \renewcommand{\arraystretch}{1}
    \renewcommand{\aboverulesep}{3pt}
    \renewcommand{\belowrulesep}{3pt}
    \setlength{\tabcolsep}{2pt}
    \fontsize{14pt}{14} \selectfont
    \resizebox{\linewidth}{!}{
    \begin{tabular}{c|c|cccc|cc}
    \toprule
    \texttt{500} Pairs & \textbf{Method} & \textbf{V-L Model (GB)} & \textbf{Expert train (h)} & \textbf{Data init (s)} & \textbf{Distillation (s)} & \textbf{Total GPUh} & \textbf{Perf.} \\
    \midrule
    \multirow{2}{*}{\texttt{F30k}} 
    & \cite{zhao2025edge} & 2.58 &  12.5 (1 SD expert) &  -  & 48,960$^{\textcolor{red}{\ast}}$ & 26.1 & 25.8 \\
    & \cellcolor{Goldenrod!20}  MDM        & \cellcolor{Goldenrod!20}  0.60 & \cellcolor{Goldenrod!20}  25.7 (20 experts) & \cellcolor{Goldenrod!20}  219 & \cellcolor{Goldenrod!20}  291      & \cellcolor{Goldenrod!20}  25.8 & \cellcolor{Goldenrod!20} 30.6 \\
    \midrule
    \multirow{2}{*}{\texttt{COCO}} 
    & \cite{zhao2025edge} & 2.58 & 49.0 (1 SD expert)   &  -  & 27,720$^{\textcolor{red}{\ast}}$  & 56.7 & 7.9 \\
    & \cellcolor{Goldenrod!20}  MDM        & \cellcolor{Goldenrod!20}  0.60 & \cellcolor{Goldenrod!20}  91.7 (20 experts)  & \cellcolor{Goldenrod!20} 953 & \cellcolor{Goldenrod!20}  2,765 & \cellcolor{Goldenrod!20} 92.7 & \cellcolor{Goldenrod!20} 15.3\\
    \bottomrule
    \end{tabular}
    }
    \vspace{-5mm}
\end{table}

\subsection{Full Retrieval Results} \label{sec:supp:full_results}
We provide complete image-text retrieval results with standard deviations in five random evaluations in Table~\ref{tab:main_full}.

\subsection{Full Algorithm} \label{sec:supp:algorithm}
We outline our multimodal distribution matching algorithm in Alg.~\ref{supp:alg:mdm}.
\begin{algorithm*}[t]
  \caption{Multimodal Distribution Matching}
  \label{supp:alg:mdm}
  \renewcommand{\algorithmicrequire}{\textbf{Input:}}
  \renewcommand{\algorithmicensure}{\textbf{Output:}}
  \begin{algorithmic}[1]
    \REQUIRE
      \;Real image-text dataset $\mathcal{D}_{\mathrm{real}}=\{(x_i,t_i)\}_{i=1}^{B}$; \\
      \qquad Unified pretrained image-text model $\Psi_0 := \{\theta^v_{0}$, $\theta^t_{0}\}$ with frozen text encoder; \\
      \qquad Buffer of finetuned experts $\mathcal{B}=\{(\theta^{v,(i)}_{real}, \theta^{t,(i)}_{real})\}_{i}^{N_{\mathrm{experts}}}$; \\
      \qquad Number of synthetic pairs $\tilde B$; Maximum distillation iterations $T_{\max}$; \\      
    \ENSURE 
      Optimized synthetic set $\mathcal{D}_{\mathrm{syn}}^{\star} = \{(\tilde x_j,\tilde t_j)\}_{j=1}^{\tilde B}$.      
    \vspace{0.4em}    
    \STATE \textcolor{cvprblue!80}{\textbf{Synthetic data initialization}}
    \STATE Compute joint real embeddings with $\Psi_0$: 
        \qquad $(z^v_i, z^t_i) = \Psi_0(x_i, t_i) \in \mathbb{R}^d\;\;\;\mathrm{for}\;\;i=1,\dots,B$
    \STATE  Run $k$-means clustering on joint embeddings with $K=\tilde B$: 
        \qquad $\{c_{k}\}_{k=1}^{\tilde B} \gets \mathtt{KMeans}(\{ z^v_{i}; z^t_{i})\}_{i}^{B})$,\\
    \STATE For each cluster $c$, select the real index $i_c$ whose joint feature $\{z^v;z^t\}_{i_c}$ is closest to the centroid
    \STATE Initialize the synthetic dataset by selecting at these indices from the real dataset:
      \quad $\mathcal{D}_{\mathrm{syn}}  \leftarrow \{(x_{i_c},\theta^t_{0}(t_{i_c}))\}_{c=1}^{\tilde B},$      
    \vspace{0.4em}           
    \WHILE{$t < T_{\max}$}
      \vspace{0.2em}      
      \STATE \textcolor{cvprblue!80}{\textbf{Model initialization}}
      \STATE Sample $N$ expert checkpoints: $\{\theta^{v,(1,\cdots,N)}_{\mathrm{real}},\theta^{t,(1,\cdots,N)}_{\mathrm{real}}\} \subset \mathcal{B}$.
      \STATE For each trainable layer $\ell$ of image encoder and text projector, 
        merge weights using Eq.~\textcolor{cvprblue}{5} (for $N=2$): \\
          \qquad 
          $
          \theta^{m}_{*,\ell} = \theta^{m}_{0,\ell} +\alpha\,t^{m}_{\ell}\cdot \tfrac{1}{2}\big(\Delta^{m}_{1,\ell}+\Delta^{m}_{2,\ell}\big),
            \;\; \mathrm{for}\; m\in\{v,t\},
          $
      \STATE Instantiate trainable image-text model
        $\Psi_t \gets (\theta^{v}_{*}, \theta^{t}_{*})$ with text encoder kept frozen
      \vspace{0.2em}
      \STATE \textcolor{cvprblue!80}{\textbf{Optimize synthetic data}}
      \STATE Sample a real minibatch: $\{(x_i,t_i)\}_{i=1}^{B_r} \subset \mathcal{D}_{\mathrm{real}}$      
      \STATE Encode joint features for real and synthetic:\\
        \qquad
        $(z^v_{r,i},z^t_{r,i}) \gets \mathtt{stop\_grad}(\Psi_{t}(x_i,\theta_{t}^t(t_i))), \quad
        (z^v_{s,i},z^t_{s,i}) \gets \Psi_{t}(\mathcal{D}_{\mathrm{syn}})$      
      \STATE Construct cross-modal agreement and discrepancy vectors after $\ell_2$ normalization: \\
        $
          u_{\mathrm{r},i} = \mathrm{norm}(z^{v}_{\mathrm{r},i} + z^{t}_{\mathrm{r},i}),\quad
          g_{\mathrm{r},i} = \mathrm{norm}(z^{v}_{\mathrm{r},i} - z^{t}_{\mathrm{r},i}), 
            \qquad
          u_{\mathrm{s},i} = \mathrm{norm}(z^{v}_{\mathrm{s},i} + z^{t}_{\mathrm{s},i}),\quad
          g_{\mathrm{s},i} = \mathrm{norm}(z^{v}_{\mathrm{s},i} - z^{t}_{\mathrm{s},i}) 
        $
      \vspace{0.2em}      
      \STATE Compute bidirectional InfoNCE loss on synthetic embeddings using Eq.~\textcolor{cvprblue}{3}:\\
        \qquad 
        $\mathcal{L}_{\mathrm{InfoNCE}} = {1 \over 2} \{ \mathcal{L}_{\mathrm{i2t}}(z^{v}_{\mathrm{s}},z^{t}_{\mathrm{s}}) + \mathcal{L}_{\mathrm{t2i}}(z^{t}_{\mathrm{s}},z^{v}_{\mathrm{s}}) \}$
      \vspace{0.2em}      
      \STATE Compute geodesic kernel energies on $u$ and $g$ using Eq.~\textcolor{cvprblue}{9}:\\
    \qquad
    $
    \mathcal{L}_{\mathrm{agr}} = \mathsf{GKE}\big(\{u_{r,i}\}_{i=1}^{B_r}, \{u_{s,i}\}_{i=1}^{\tilde B}\big),\quad
    \mathcal{L}_{\mathrm{dis}} = \mathsf{GKE}\big(\{g_{r,i}\}_{i=1}^{B_r}, \{g_{s,i}\}_{i=1}^{\tilde B}\big),
    $
      \vspace{0.2em}      
      \STATE Update synthetic data by the total loss objective using Eq.~\textcolor{cvprblue}{11}: \\
      \qquad
        $
          \mathcal{L}_{\mathrm{MDM}} = \mathcal{L}_{\mathrm{InfoNCE}} + \lambda_{\mathrm{agr}}\cdot\mathcal{L}_{\mathrm{agr}} + \lambda_{\mathrm{dis}} \cdot\mathcal{L}_{\mathrm{dis}}.
        $
      \STATE Update only the synthetic parameters by gradient descent:
        $
          \mathcal{D}_{\mathrm{syn}} \gets \mathcal{D}_{\mathrm{syn}} - \eta\,\nabla_{\mathcal{D}_{\mathrm{syn}}} \mathcal{L}_{\mathrm{MDM}}
        $
      \vspace{0.2em}      
      \STATE $t \gets t + 1$
    \ENDWHILE
    \STATE \textbf{return} $\mathcal{D}_{\mathrm{syn}}^{\star}$

  \end{algorithmic}
\end{algorithm*}

\end{document}